\pdfoutput=1
\documentclass[11pt]{article}

\usepackage[final]{acl}

\usepackage{times}
\usepackage{latexsym}

\usepackage[T1]{fontenc}

\usepackage[utf8]{inputenc}

\usepackage{microtype}

\usepackage{inconsolata}

\usepackage{graphicx}

 \usepackage[table]{xcolor}

\usepackage{amsmath}
\usepackage{amssymb}

\usepackage{booktabs}
\usepackage{multirow}
\usepackage{array}

\usepackage{enumitem}

\usepackage{titletoc}

\title{\textsc{EviGen}: Predictive Evidence Scaffolding for Verifiable Clinical Rationale Generation}

\author{
  \textbf{Fengnan Li}\textsuperscript{*},
  \textbf{Heman Burre}\textsuperscript{*},
  \textbf{Liwen Sun},
  \textbf{Roshni Varma},
  \textbf{Matthew M. Engelhard}
\\
  Duke University
\\
  \texttt{\{fengnan.li, heman.burre, l.sun, roshni.varma, m.engelhard\}@duke.edu}
}

\begin{document}
\maketitle
\begingroup
\renewcommand\thefootnote{}\footnotetext{\textsuperscript{*}\,Equal contribution.}
\endgroup
\begin{abstract}

Longitudinal electronic health records (EHRs) capture years of patient history across notes, codes, labs, and procedures, and contain evidence needed to reason about likely clinical outcomes. However, comprehensive clinician review of these records is impractical, and LLM-based processing is costly and often unreliable, missing some relevant observations while hallucinating others. We therefore propose \textsc{EviGen}, a three-layer framework for verifiable clinical rationale generation that addresses these challenges. The \textbf{first layer} is a patient-conditioned retriever that uses learnable queries to find evidence \emph{predictive} of, not just textually relevant to, a clinical outcome and ranks it by prediction attribution scores. The \textbf{second layer} is an LLM generator that consumes this ranked evidence as a scaffold to produce a clinical rationale grounded in the retrieved spans. The \textbf{third layer} is a process-supervised verifier that checks the generated rationale at the reasoning-step level, flagging unreliable claims. Across three medical prediction datasets, \textsc{EviGen} improves prediction performance and rationale faithfulness over full-context LLM and RAG baselines, and is preferred by clinical reviewers in a usability evaluation.\footnote{Code is available at \url{https://github.com/engelhard-lab/EviGen}}
\end{abstract}

\section{Introduction}

\begin{figure}[!t]
  \centering
  \includegraphics[width=\columnwidth]{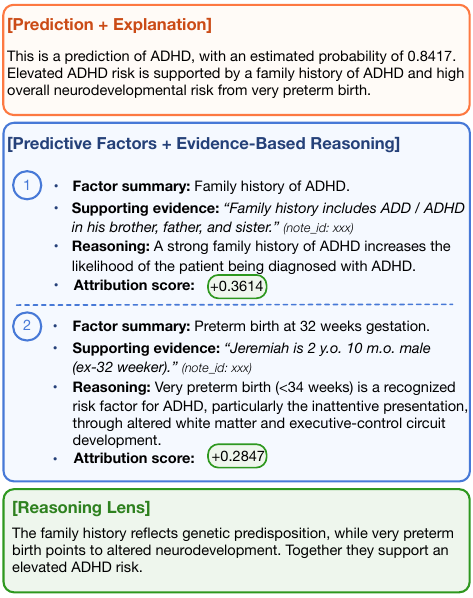}
  \caption{An \textsc{EviGen}-generated clinical rationale, simplified and de-identified for illustration. Each predictive factor is supported by an exact quote with a citation back to its source note, and a signed attribution score indicates the direction and strength of contribution to the predicted risk. Every claim is traceable to the original record, and a process-supervised verifier (\S\ref{sec:method:verification}) will then flag unreliable reasoning steps for clinician review (full example in Figure~\ref{fig:flagged-example}).}
  \label{fig:report-example}
\end{figure}

Electronic health records (EHRs) contain rich information for clinical assessment, diagnosis, and risk prediction \citep{mohsen2022artificial,li2022integrating,engelhard2023predictive}. A patient's record often spans years of care and includes diagnoses, laboratory results, procedures, medications, and free-text notes. While some clinical decisions can be made from a single encounter, others depend on sparse evidence distributed across the record \citep{kruse-etal-2025-large}. For example, in pediatric ADHD assessment (Figure~\ref{fig:flagged-example}), relevant evidence includes birth encounter details such as prematurity, visits to speech and vision specialists that reflect early developmental concerns \citep{engelhard2020util}, and longitudinal observations recorded by pediatricians across routine well-child visits. The ADHD signal emerges only when temporally distant and heterogeneous pieces of evidence are integrated, making longitudinal EHR reasoning difficult for both clinicians and automated systems \citep{dymek2021building}.

Large language models (LLMs) provide a promising approach to processing clinical records, including summarization, diagnostic reasoning, and recommendation generation \citep{gao-etal-2023-overview,van2024adapted,williams2024evaluating,goh2024large}. However, applying an LLM directly to a patient's full EHR is unreliable: records may exceed the model's effective context length, and important evidence can be ignored when it is buried in a long input \citep{liu-etal-2024-lost}. Even when records fit, the quadratic cost of attention makes LLM inference over full patient histories impractical at scale. %

Retrieval-augmented generation (RAG) is the standard solution to this challenge: it reduces long inputs by selecting context before generation \citep{lewis2020retrieval} and has been widely studied in clinical NLP \citep{xiong-etal-2024-benchmarking,wu2024medical,lopez2025clinical}. However, effective retrieval remains a bottleneck for longitudinal clinical reasoning. Standard RAG retrieves text that is relevant to an explicit query, but clinically important evidence is often indirect, sparsely distributed, and dependent on encounters across the EHR \citep{li-etal-2025-iris}. Because the same outcome can arise through many different clinical pathways, query-based retrieval must either pre-specify those pathways, risking missed evidence, or retrieve broadly, adding weakly relevant context \citep{sohn-etal-2025-rationale}. %

Beyond retrieval quality, hallucination further limits the reliability of clinical LLMs. Unsupported diagnoses, fabricated events, and unjustified recommendations are documented failure modes in medical text generation \citep{asgari2025creola}. In clinical settings, such errors substantially undermine clinician trust, as providers remain responsible for decisions informed by the model’s output. This challenge is amplified for longitudinal EHRs: because patient histories can be long, it is impractical for clinicians to audit every generated claim against the original record. As a result, clinical trust requires not only accurate predictions, but reasoning that is explicitly grounded in auditable and traceable patient evidence.

To address these limitations, we argue that LLM reasoning over longitudinal EHRs should not proceed directly from the full record or from generic retrieved context. Instead, the system should first build a \emph{predictive evidence scaffold}: a compact set of observations selected for \emph{predictive} contribution, not textual relevance, to the target task. The LLM can then reason over this scaffold, with each reasoning step grounded in selected evidence. This grounding directly supports model \emph{faithfulness}: that generated claims are supported by evidence in the patient's record. More importantly, it enables clinicians to inspect the specific evidence underlying each prediction and reasoning step, rather than relying on black box rationales. In high-stakes clinical settings, such traceability is critical for establishing providers' trust, as they will be able to verify the model's reasoning before incorporating its conclusion into patient care.

We propose \textsc{EviGen}, a three-layer framework for verifiable clinical rationale generation. The \textbf{Evidence Selection Layer} learns patient-conditioned queries that identify evidence predictive of the target task, produces a prediction, and ranks each piece of selected evidence by its contribution to that prediction. The \textbf{Rationale Generation Layer} prompts an LLM to generate a structured rationale over this ranked evidence. Each reasoning step must quote a supporting passage in the evidence and link back to the EHR through a citation ID (Figure~\ref{fig:report-example}). The \textbf{Process Verification Layer} verifies each reasoning step and flags unreliable ones for clinician review. Because every step is tied to cited evidence, clinicians can trace flagged errors back to the source. Together, these layers connect prediction, rationale generation, and verification through the same patient-specific evidence.

Our contributions are as follows:
\begin{itemize}
  \item We introduce \textsc{EviGen}, to our knowledge the first framework for clinical rationale generation over full longitudinal EHR histories spanning years of patient care.

  \item We introduce learnable queries as evidence detectors that retrieve heterogeneous evidence (e.g., clinical notes and structured codes), selectively activated per patient to enable efficient retrieval over arbitrarily long input.

  \item We generate rationales through verifiable evidence-scaffolded reasoning: each reasoning step is based on selected evidence, traceable to the original record, auditable by a verifier, and designed to support clinician trust through transparent evidence attribution.

  \item Across three longitudinal clinical tasks, \textsc{EviGen} substantially improves prediction performance and rationale faithfulness over full-context LLM and RAG baselines, and is preferred by clinical reviewers in a usability pilot study.
\end{itemize}

\section{Related Work}

\paragraph{Query-Based Context Retrieval.} Standard RAG retrieves text similar to a given query \citep{lewis2020retrieval}. Prior work has improved retrieval with query expansion, rewriting, and iterative refinement in medical contexts \citep{wang2023query2doc, ma2023query, xiong2024improving}. In all cases, queries retrieve generally relevant context, but may miss the nuanced, patient-specific evidence required for adequate clinical reasoning \citep{sohn-etal-2025-rationale}.

Closest to our setting, \textsc{IRIS} \citep{li-etal-2025-iris} uses learnable query vectors trained on outcome labels. Unlike RAG, its retrieval is driven by predictive utility rather than textual similarity, with each query learning to capture a risk factor. \textsc{EviGen} builds on this idea in two ways. First, it introduces patient-conditioned query activation, so that only the subset of queries relevant to a patient's clinical profile is used, enabling more tailored, patient-specific retrieval. Second, it extends predictive evidence retrieval beyond document classification by using the selected observations to guide generation.

\paragraph{Process-Supervised Verification.} Process supervision scores intermediate reasoning steps rather than the final answer alone \citep{lightman2024let}. Recent work \citep{wang-etal-2025-process} applies this paradigm to verify AI-generated clinical notes, breaking each note into clinically meaningful steps and checking each for reasoning and factual issues. \textsc{EviGen} adapts this step-level verification approach to audit generated clinical rationales. It outputs type-aware error flags (e.g., hallucination, factual inaccuracy) at unreliable reasoning steps, and clinicians can trace each flag back to the cited observation via citation IDs.

\section{\textsc{EviGen}}
\label{sec:method}

\subsection{Overview}
\label{sec:method:overview}

\paragraph{Task.} Given a patient's longitudinal EHR, \textsc{EviGen} produces: (i) a risk prediction probability $\hat{y}$ and (ii) a structured clinical rationale that traces every claim to a passage in the patient's record.

\begin{figure*}[t]
  \centering
   \includegraphics[width=\textwidth]{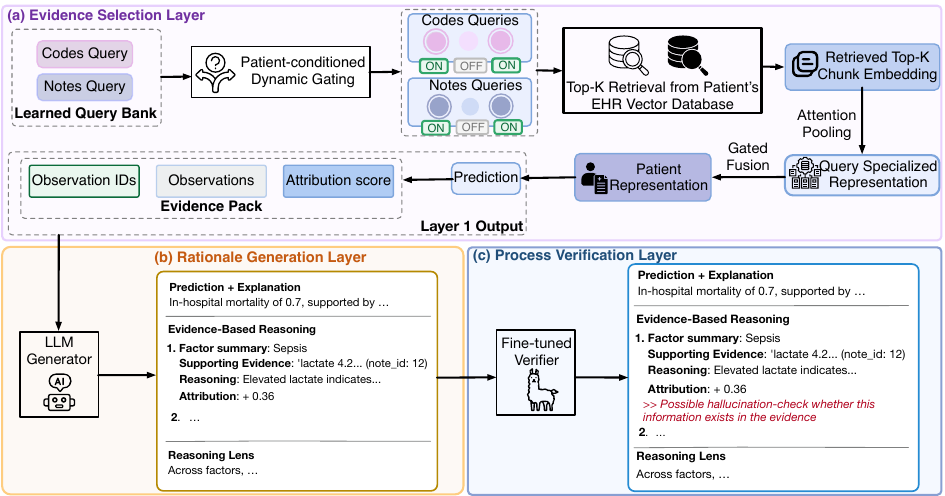}
  \caption{\textbf{\textsc{EviGen} three-layer pipeline.} (a) Evidence Selection Layer (\S\ref{sec:method:evidence}); (b) Rationale
  Generation Layer (\S\ref{sec:method:generation}); (c) Process Verification Layer (\S\ref{sec:method:verification}).}
  \label{fig:pipeline}
\end{figure*}

\paragraph{Framework.} As shown in Figure~\ref{fig:pipeline}, \textsc{EviGen} consists of three layers with explicit input-output interfaces. The \textbf{Evidence Selection Layer} (\S\ref{sec:method:evidence}) takes the full EHR, extracts the predictive observations, and outputs a prediction probability $\hat{y}$ with a ranked evidence pack containing each observation's ID and signed attribution score. The \textbf{Rationale Generation Layer} (\S\ref{sec:method:generation}) consumes the probability $\hat{y}$ and the evidence pack, generating a structured clinical rationale whose claims cite the provided observation IDs. Finally, the \textbf{Process Verification Layer} (\S\ref{sec:method:verification}) takes the generated rationale as input, reviews each reasoning step, and flags steps with potential reasoning errors. Observation IDs are preserved across each layer, making the final output traceable to the source record.

\paragraph{Input.} We use two EHR modalities: free-text clinical notes (split into chunks) and structured ICD diagnostic codes (encoded from their official descriptions). Both are embedded with a shared text encoder, producing note embeddings $\mathbf{X}^{(n)} = \{\mathbf{x}^{(n)}_\ell\}_{\ell=1}^{L_n}$ and code embeddings $\mathbf{X}^{(c)} = \{\mathbf{x}^{(c)}_\ell\}_{\ell=1}^{L_c}$, with $\mathbf{x}^{(m)}_\ell \in \mathbb{R}^d$, for a given patient.

\subsection{Evidence Selection Layer}
\label{sec:method:evidence}

\paragraph{Overview.} This layer identifies which observations in a patient's EHR (\textit{i.e.}, note chunks and ICD codes) are most predictive of the clinical outcome. It outputs (i) a risk prediction probability $\hat{y}$ and (ii) an evidence pack: a set of (observation, attribution) tuples, where each top-attributed observation carries a unique ID and a signed attribution score, passed to the Rationale Generation Layer.

The challenge in this layer is to retrieve predictive signals from an extensive EHR input. Standard RAG queries cannot capture the heterogeneous, patient-specific evidence patterns present in longitudinal EHRs. Instead, this layer uses a set of learnable query vectors $\mathbf{Q} = [\mathbf{q}_0, \dots, \mathbf{q}_{N-1}] \in \mathbb{R}^{N \times d}$, partitioned into $N_n$ note queries (retrieving note chunks) and $N_c$ code queries (retrieving ICD codes). Queries are trained end-to-end with the prediction objective using clinical outcome labels, so each query specializes in detecting a distinct evidence type predictive of the outcome. Outcome labels such as diagnoses or adverse events can be obtained from structured EHR fields via established computable phenotypes \citep{kirby2016phekb}, enabling large-scale supervised training of query vectors without human annotations. Since not all evidence types are present in every patient, we further introduce a gating mechanism that activates only the queries appropriate for a given patient.

\paragraph{Patient-conditioned dynamic gating.} To make query activation dependent on a patient's specific profile, we compute a patient-specific global summary $\mathbf{s}^{(m)} \in \mathbb{R}^d$ for each modality $m$ via mean pooling over all chunk embeddings for that modality ($\mathbf{X}^{(n)}$ for notes, $\mathbf{X}^{(c)}$ for codes). Query $\mathbf{q}_i$ fires for modality $m$ only when its alignment with this summary exceeds a threshold $\eta_i$ (i.e., the evidence type targeted by $\mathbf{q}_i$ is reflected in the summary):
\begin{equation}
g_i^{(m)} = \mathbf{1}\!\left[\,\sigma\!\left(\tilde{\mathbf{s}}^{(m)\top}\tilde{\mathbf{q}}_i\right) > \sigma(\eta_i)\,\right]
\label{eq:gating}
\end{equation}
Here $\tilde{\cdot}$ denotes L2 normalization, so $\tilde{\mathbf{s}}^{(m)\top}\tilde{\mathbf{q}}_i$ is the cosine similarity between the summary and the query; $\eta_i \in \mathbb{R}$ is a learnable per-query threshold; and $g_i^{(m)} \in \{0, 1\}$ is the binary gate indicating whether query $i$ activates for modality $m$. Since the indicator $\mathbf{1}[\cdot]$ is non-differentiable, we apply a straight-through estimator \citep{bengio2013estimating} (Appendix~\ref{app:a1:ste}) so $\mathbf{q}_i$ and $\eta_i$ remain trainable.

\paragraph{Retrieval and attention pooling.} After gating, for each active query $\mathbf{q}_i$ in modality $m$, we retrieve the top-$K$ embeddings from $\mathbf{X}^{(m)}$ by cosine similarity to $\mathbf{q}_i$ and pool the retrieved embeddings via the attention pooling approach used by \textsc{IRIS} \citep{li-etal-2025-iris}:
\begin{equation}
\mathbf{c}_i = \sum_{j} \alpha_{i,j}\, \mathbf{x}_j^{(m)}, \quad \alpha_{i,j} = \tfrac{\exp(\tilde{\mathbf{q}}_i^\top \mathbf{x}_j^{(m)})}{\sum_{j'} \exp(\tilde{\mathbf{q}}_i^\top \mathbf{x}_{j'}^{(m)})}
\label{eq:attention}
\end{equation}
In words: $\mathbf{c}_i$ is a weighted average of the top-$K$ retrieved chunks, where chunks more similar to query $i$ receive more weight.
Here $\mathbf{x}_j^{(m)}$ is the $j$-th retrieved embedding from modality $m$, $\alpha_{i,j}$ is its softmax attention weight, and $\mathbf{c}_i \in \mathbb{R}^d$ is the pooled context vector summarizing query $i$'s observations. Although top-$K$ hard retrieval is non-differentiable, the queries remain trainable through these attention weights, and during training, they converge to retrieve predictive chunks. We discuss this learning dynamic in Appendix~\ref{app:a1:retrieval}.

The pooled context $\mathbf{c}_i$ for each active query is then refined by its corresponding per-query expert MLP, producing a query-specialized representation $\mathbf{h}_i$. Because dynamic gating restricts each expert MLP to patients whose histories match its corresponding query, each MLP can specialize to a narrowly defined predictive pattern rather than a broad set of predictive content. Finally, all $\{\mathbf{h}_i\}$ from active queries are aggregated into a single patient embedding $\bar{\mathbf{h}}$, which is forwarded to the prediction head to produce $\hat{y}$.

\paragraph{Training and inference.} Query vectors, per-query gating thresholds, per-query MLPs, and the prediction head are trained end-to-end with cross-entropy loss and a diversity regularizer \citep{guo2025dynamic} that prevents different queries from converging onto the same pattern (Appendix~\ref{app:a1:training}). At inference, we use the trained queries to retrieve predictive observations and apply Integrated Gradients \citep{sundararajan2017axiomatic} to the trained predictor to obtain a signed attribution score $\alpha_i$ for each retrieved chunk, quantifying its contribution to the risk prediction probability, $\hat{y}$. The top-attributed chunks, together with their observation IDs and attribution scores, constitute the \emph{evidence pack} forwarded to Layer~2.

\subsection{Rationale Generation Layer}
\label{sec:method:generation}

\paragraph{Overview.} The Rationale Generation Layer generates a clinical rationale from the Evidence Selection Layer's evidence pack and prediction probability $\hat{y}$. The rationale is constrained to ground every claim in the selected evidence and to surface each observation's signed attribution score alongside its reasoning.

\paragraph{Rationale structure.} The rationale structure is illustrated in Figure~\ref{fig:report-example}. For each observation in the evidence pack, an LLM is prompted to write a structured factor section with four fields per factor: a factor summary, a verbatim quote with a citation ID tracing back to the original record, brief reasoning over the quoted observation, and an attribution score. Afterwards, the LLM writes a reasoning lens section that synthesizes the factor-level reasoning into a coherent analysis, highlighting any interactions between factors.

\paragraph{Rationale verifiability.} \textsc{EviGen}'s LLM generator may still misquote or paraphrase the source despite receiving the evidence pack. We therefore use string matching (Appendix~\ref{app:faithfulness:stage1}) to verify that every quoted observation appears in the source record it cites; mismatches are flagged. With every quote verified and every reasoning step grounded solely in those quotes, the rationale itself becomes auditable: a clinician can review it without reading the entire EHR. %
In contrast, standard LLM explanations require clinicians to verify claims against the full input, which is often impractical.

\subsection{Process Verification Layer}
\label{sec:method:verification}

\paragraph{Overview.} While \textsc{EviGen} rationales are verifiable, clinicians face limited time and high cognitive burden \citep{dymek2021building, rule2021length}, and they may miss errors in the rationale. To support clinician trust and direct attention to potentially unreliable claims, this layer flags suspicious reasoning in generated rationales for review.

\paragraph{Process-supervised verifier.} Following the process supervision paradigm of \citet{lightman2024let}, we train a step-level verifier to screen each generated rationale for reasoning and factual errors. The training approach adapts \citet{wang-etal-2025-process}: we fine-tune Llama-3.1-8B-Instruct on synthetic negative samples produced by prompting GPT-4o-mini to perturb individual steps according to one of five error types. After each reasoning step, the verifier outputs a softmax over special reserved tokens (one for "correct" and one per error type), giving per-step probabilities for each outcome. At inference, per-step probabilities are Platt-calibrated, and steps below the calibrated threshold are flagged, with the predicted error type attached (Figure~\ref{fig:flagged-example}). The error type descriptions, full training recipe, and verifier performance metrics are in Appendix~\ref{app:verification-layer}.

\section{Experimental Setup}
\label{sec:experiments}

Our experiments address three research questions. \textit{RQ1:} Does \textsc{EviGen} improve prediction performance? \textit{RQ2:} Does \textsc{EviGen} generate more faithful rationales? \textit{RQ3:} Do clinical reviewers find \textsc{EviGen} rationales useful for clinical decision making? This section discusses datasets, baselines, and evaluation methods to answer these questions.

\subsection{Datasets}

Our method is designed for longitudinal histories, but few public EHR datasets contain comprehensive longitudinal clinical notes due to privacy constraints. As a partial substitute, we use discharge notes from \textbf{MIMIC-IV}~\citep{johnson2023mimic}, selecting patients with rich discharge history and predicting 1-year all-cause mortality from their last hospital discharge. Since other (non-discharge) notes are not available, the input is short and information-dense, making evidence retrieval easier than in real longitudinal EHR settings.

To complement MIMIC-IV with longer and sparser records, we also focus on early prediction of (a) \textbf{autism spectrum disorder (ASD)} and (b) \textbf{attention-deficit/hyperactivity disorder (ADHD)} from complete longitudinal EHRs from our institution. For each patient, we use the full sequence of outpatient visits (e.g., well-child checks, sick visits, specialty referrals) up to age 1.5 for autism and age 3 for ADHD, both well prior to the typical age of clinical diagnosis \citep{loh2025limitations}.

The MIMIC-IV mortality dataset contains $\sim$5.5K positive and $\sim$8K negative patients, averaging $\sim$14K input tokens per patient. The autism and ADHD datasets are smaller in case count ($\sim$1.6K and $\sim$1.7K positives, each paired with 8K negatives) but substantially longer in input, averaging $\sim$101K and $\sim$108K tokens per patient, respectively. Inclusion/exclusion criteria and additional statistics are in Appendix~\ref{app:datasets}.

All three datasets are split 8:1:1 into train, validation, and test partitions. \textsc{EviGen} is trained on the train set with model selection on the validation set, and all methods (\textsc{EviGen} and baselines) are evaluated on the test set.

\subsection{Baselines}
\label{sec:exp:baselines}

We pair four LLMs with two input setups and two prediction strategies, yielding 16 baselines per task.

\paragraph{LLM backbones.} Llama-3.1-8B/70B-Instruct, Qwen3-32B (thinking mode), and GPT-4o-mini, all run with a 128K-token context window.

\paragraph{Input setups.} \emph{Full-context zero-shot} places the full patient record directly in the prompt. \emph{Retrieval-augmented generation (RAG)} performs dense retrieval over the patient record using multi-factor queries; the queries are per-task risk factors compiled by Claude Opus 4.7 from the clinical literature and verified by clinical experts.

\paragraph{Prediction strategies.} \emph{Free-form} asks the model to output a probability directly. \emph{Yes/No verbalizer} \citep{schick-schutze-2021-exploiting} asks for a Yes/No answer and reads the normalized probability over the Yes and No tokens. All baselines are prompted to generate a rationale in the same format as \textsc{EviGen} (Figure~\ref{fig:report-example}), without an attribution score, as these baselines do not calculate such a score. Implementation details are in Appendix~\ref{app:baselines}.

\subsection{Evaluation}
\paragraph{Prediction Performance.} We report accuracy and AUC on MIMIC-IV mortality, and AUC only on the autism and ADHD datasets, as the latter two are heavily class-imbalanced and accuracy is less informative.

\paragraph{Rationale Faithfulness.}\label{sec:eval:faithfulness} With inputs averaging $\sim$100K tokens, clinician review is impractical. LLM-as-a-judge \citep{chiang2023can} is the standard alternative, but judge quality degrades at this scale.

Our evaluation design exploits the rationale format: each factor contains a citation ID, a quoted observation, and a brief reasoning over that quote (\S\ref{sec:method:generation}). The quote sits between input and reasoning, splitting the chain \textit{input} $\rightarrow$ \textit{reasoning} into two sub-checks: \textit{input} $\rightarrow$ \textit{observation} (does the quote exist in the EHR?) and \textit{observation} $\rightarrow$ \textit{reasoning} (does the reasoning follow from the quote?). Stage 1 reduces to string matching and needs no LLM; Stage 2 passes only the rationale to an LLM judge.

\textbf{Stage 1: source grounding.} For each factor, we verify that (i) the citation ID points to a real ID in the patient's record, and (ii) the quoted observation matches a contiguous span of the cited source note. The matching check is based on the longest-common-subsequence algorithm.

\textbf{Stage 2: reasoning consistency.} We pass only the rationale (not the full EHR) to GPT-4o. For each factor, the LLM judge checks whether (i) the factor's reasoning follows from its quoted observation, and (ii) the reasoning introduces no facts beyond the observation. For the overall reasoning lens section, the judge additionally checks whether (iii) the lens follows from the per-factor reasoning, and (iv) the lens relies only on the stated factors.

For a fair comparison, \textsc{EviGen} and each baseline produce exactly five factors per rationale, yielding 22 checks per rationale (4 per factor + 2 overall). A rationale is considered faithful only if it passes all 22 checks. Implementation is detailed in Appendix~\ref{app:faithfulness}.

\paragraph{Clinical Utility.} We conduct a pilot evaluation to assess \textsc{EviGen} rationales' clinical utility. We ask 7 medical students to review 3 types of rationales: (i) the \textsc{EviGen} rationale, (ii) a standard LLM clinical rationale (prediction with a free-form explanation), and (iii) the raw output of \textsc{EviGen}'s Evidence Selection Layer (prediction probability with the top-attribution chunks and their scores, no LLM-generated narrative). Reviewers filled out surveys to comprehensively rate each rationale type. Questions in the survey are drawn from validated user-evaluation scales for clinical decision support tools \citep{brooke1996sus, holzinger2020measuring, ghorayeb2023design, hoffman2023measures}. The full survey is in Appendix~\ref{app:clinical-survey}.

\section{Results and Analysis}
\label{sec:results}

\subsection{Prediction Performance}
\label{sec:results:prediction}
\textsc{EviGen} outperforms all baselines on all four metrics. The two largest performance gaps between \textsc{EviGen} and the best baseline are on MIMIC-IV mortality accuracy (+10.4pp) and ADHD AUC (+8.9pp); the performance gaps for the mortality AUC and the autism AUC are around 4pp. %

Across baseline backbones, RAG does not reliably beat full-context, suggesting that standard query-based retrieval offers limited assistance to the model's clinical reasoning. In contrast, \textsc{EviGen} retrieves observations using queries trained on outcome labels, so the selected context is optimized for predictive utility rather than textual relevance. Baseline AUCs show some discrimination ability, but their low mortality accuracy reflects poor calibration: Llama-3.1-8B/70B and GPT-4o-mini all post accuracies below 0.5 despite AUCs of 0.69–0.80. On the mortality task, \textsc{EviGen} achieves an Expected Calibration Error (ECE) of 0.042, compared with 0.200 to 0.536 for the zero-shot and RAG baselines. \textsc{EviGen}'s combination of high accuracy (0.800), high AUC (0.871), and low ECE supports both discrimination and well-calibrated prediction.

\S\ref{sec:exp:baselines} introduced two prediction strategies for baselines (free-form and verbalizer); we find no pattern across model-task pairs as to which strategy outperforms the other, and Tables~\ref{tab:prediction} and~\ref{tab:faithfulness} report the better of the two per baseline.

\begin{table}[t]
\centering
\small
\setlength{\tabcolsep}{6pt}
\begin{tabular}{l r r @{\hskip 6pt} r @{\hskip 6pt} r}
\toprule
 & \multicolumn{2}{c}{Mortality} & Autism & ADHD \\
\cmidrule(lr){2-3}
\textbf{Method} & \textit{Acc.} & \textit{AUC} & \textit{AUC} & \textit{AUC} \\
\midrule
\multicolumn{5}{l}{\emph{Llama-3.1-8B}} \\
\quad Full context & 0.446 & 0.694 & 0.526 & 0.558 \\
\quad RAG          & 0.448 & 0.782 & 0.635 & 0.583 \\
\cmidrule(lr){1-5}
\multicolumn{5}{l}{\emph{Llama-3.1-70B}} \\
\quad Full context & 0.419 & 0.799 & 0.657 & 0.612 \\
\quad RAG          & 0.414 & 0.799 & 0.667 & 0.598 \\
\cmidrule(lr){1-5}
\multicolumn{5}{l}{\emph{Qwen3-32B}} \\
\quad Full context & 0.690 & 0.831 & 0.687 & 0.607 \\
\quad RAG          & 0.696 & 0.811 & 0.691 & 0.588 \\
\cmidrule(lr){1-5}
\multicolumn{5}{l}{\emph{GPT-4o-mini}} \\
\quad Full context & 0.407 & 0.710 & 0.679 & 0.615 \\
\quad RAG          & 0.478 & 0.772 & 0.654 & 0.594 \\
\addlinespace[2pt]
\midrule
\textbf{\textsc{EviGen} (ours)} & \textbf{0.800} & \textbf{0.871} & \textbf{0.730} & \textbf{0.704} \\
\bottomrule
\end{tabular}
\caption{Prediction performance of \textsc{EviGen} and baselines across three datasets. The best result in each column is shown in \textbf{bold}.}
\label{tab:prediction}
\end{table}

  \begin{table*}[t]
  \centering
  \footnotesize
  \setlength{\tabcolsep}{3pt}
  \begin{tabular}{ll @{\hspace{6pt}} ccc c ccc c ccc c ccc}
  \toprule
   & & \multicolumn{3}{c}{Llama-3.1-8B} & & \multicolumn{3}{c}{Llama-3.1-70B} & & \multicolumn{3}{c}{Qwen3-32B} & & \multicolumn{3}{c}{GPT-4o-mini} \\
  \cmidrule(lr){3-5}\cmidrule(lr){7-9}\cmidrule(lr){11-13}\cmidrule(lr){15-17}
  \textbf{Dataset} & \textbf{Metric} & Full & RAG & \textbf{Ours} & & Full & RAG & \textbf{Ours} & & Full & RAG & \textbf{Ours} & & Full & RAG & \textbf{Ours} \\
  \midrule
  \multirow{3}{*}{Mortality}
    & S1  & 9.0  & 18.5 & 72.6 & & 39.6 & 68.4 & 67.7 & & 28.5 & 46.2 & 55.7 & & 24.9 & 34.3 & 72.9 \\
    & S2  & 67.5 & 74.2 & 81.0 & & 80.9 & 78.5 & 83.1 & & 94.0 & 93.5 & 94.2 & & 92.6 & 93.4 & 88.6 \\
  \rowcolor{gray!15}
    & S1$\And$S2 & 6.1  & 13.7 & \textbf{58.8} & & 32.0 & 53.7 & \textbf{56.3} & & 26.8 & 43.2 & \textbf{52.5} & & 23.1 & 32.0 & \textbf{64.6} \\
  \midrule
  \multirow{3}{*}{Autism}
    & S1  & 1.8  & 30.1 & 77.9 & & 43.2 & 78.5 & 85.7 & & 6.2  & 46.3 & 49.0 & & 18.7 & 58.3 & 75.2 \\
    & S2  & 58.8 & 41.5 & 53.8 & & 77.2 & 81.5 & 75.3 & & 98.3 & 99.6 & 98.3 & & 95.5 & 95.7 & 90.9 \\
  \rowcolor{gray!15}
    & S1$\And$S2 & 1.0  & 12.5 & \textbf{41.9} & & 33.4 & 64.0 & \textbf{64.6} & & 6.0  & 46.0 & \textbf{48.1} & & 17.8 & 55.8 & \textbf{68.3} \\
  \midrule
  \multirow{3}{*}{ADHD}
    & S1  & 0.8  & 35.3 & 70.1 & & 30.3 & 78.3 & 83.0 & & 3.3  & 47.5 & 59.7 & & 8.9  & 73.1 & 73.1 \\
    & S2  & 42.9 & 40.4 & 38.0 & & 72.4 & 67.8 & 78.8 & & 100.0 & 99.8 & 99.8 & & 92.0 & 94.1 & 83.3 \\
  \rowcolor{gray!15}
    & S1$\And$S2 & 0.4  & 14.2 & \textbf{26.7} & & 21.9 & 53.0 & \textbf{65.4} & & 3.3  & 47.4 & \textbf{59.6} & & 8.2  & \textbf{68.7} & 60.8 \\
  \bottomrule
  \end{tabular}
  \caption{Rationale faithfulness on the three datasets. S1 = source grounding pass rate; S2 = reasoning consistency pass rate; S1$\And$S2 = joint pass rate across all 22 checks
  (\S\ref{sec:eval:faithfulness}). "Ours" = \textsc{EviGen}. S1$\And$S2 rows are shaded to mark the headline metric. The highest S1$\And$S2 per backbone-dataset cell is in
  \textbf{bold}.}
  \label{tab:faithfulness}
  \end{table*}

\subsection{Rationale Faithfulness}
\label{sec:results:faithfulness}

Table~\ref{tab:faithfulness} reports the pass rate of each stage (S1: source grounding, S2: reasoning consistency) and their joint pass rate (S1$\And$S2). \textsc{EviGen} achieves highest S1$\And$S2 on 11/12 backbone-dataset cells, with 0.6-45.1pp margins over the best baseline.

LLMs are more susceptible to hallucination as input length grows. RAG substantially outperforms full-context generation across all tasks and backbones. \textsc{EviGen} narrows the input further by passing only predictive evidence, rather than RAG's pool of textually relevant candidates. Because the evidence pack from the Evidence Selection Layer contains data that is already extracted and ranked, \textsc{EviGen} does not need to judge which pieces matter; it only needs to quote them. This simpler task yields less hallucination.

The smaller model, Llama-3.1-8B, substantially lags behind other backbones under full-context and RAG. \textsc{EviGen} closes this gap. For mortality, Llama-3.1-8B + \textsc{EviGen} reaches a 58.8\% joint pass rate, up from 6.1\% under full context and 13.7\% under RAG, and even exceeds Llama-3.1-70B + \textsc{EviGen} (56.3\%). The evidence scaffold for generation makes a smaller model comparable to a larger one for producing faithful rationales.

Stage 1 is the harder check for baselines, as producing exact verbatim quotes from a long context is difficult. EHR notes make this more difficult: they heavily use abbreviations and specialized terms \cite{van2024adapted}, a style rare in pre-training data and often suppressed in post-training for readability. However, verbatim quoting is important because it preserves the transparency and traceability required in clinical settings. \textsc{EviGen} substantially exceeds baselines in Stage 1.

Stage 2 pass rates are similar across methods. This is partly a survivor bias: only quotes that pass Stage 1 reach Stage 2. For baselines, the observations judged at Stage 2 are thus the more readable ones, since hard-to-quote notes were already filtered out in Stage 1. Reasoning consistency is naturally easier over clearer observations.

The only reasoning model in our set, Qwen3-32B, has a low Stage 1 pass rate but a near-perfect Stage 2 pass rate. This is consistent with recent findings that chain-of-thought can worsen instruction following \citep{qin2026incentivizing}. Qwen3-32B trades quote fidelity for superior reasoning quality. \textsc{EviGen} avoids this trade-off by providing reasoning support via the evidence scaffold, so non-reasoning backbones achieve both high S1 and S2.

\subsection{Clinical Utility}
\label{sec:results:clinical}

Our clinical reviewer pilot (Appendix~\ref{app:reviewer-pilot}) finds that \textsc{EviGen} rationales receive an average rating of 3.49/5 across all questions, compared with 3.34 for the standard LLM rationale (prediction + explanation) and 2.71 for the prediction with evidence pack alone (no LLM narrative). Six of seven reviewers selected \textsc{EviGen} as their preferred method. \textsc{EviGen} leads on informativeness, actionability, and trust, but falls short of the standard LLM rationale on usability. One likely explanation is that the simpler prediction + explanation format is easier to use, whereas \textsc{EviGen}'s richer output (attribution scores, process verifier) imposes a higher interpretive burden on medical students without an ML background.

Because reviewers cannot verify against the true EHR input, they must assume that predictions are correct and cited quotes are not hallucinated when rating each rationale. This assumption does not hold in live clinical use, and it is exactly what \textsc{EviGen}'s lead on prediction accuracy (\S\ref{sec:results:prediction}) and rationale faithfulness (\S\ref{sec:results:faithfulness}) provides.

\subsection{Additional Baseline Comparisons}
\label{sec:ablation}
\textsc{EviGen}'s gain over zero-shot baselines could reflect access to training labels rather than more effective sparse-evidence selection. To control for this, we fine-tune baselines on the same train and validation set as \textsc{EviGen} and compare. We fine-tune Llama-3.1-8B-Instruct with QLoRA (4-bit base, rank 16) with two input settings: full context (each patient's record is truncated to its most recent 32K tokens due to the compute burden of long inputs) and RAG-retrieved input. Both settings use the Yes/No verbalizer (§\ref{sec:exp:baselines}). We also include \textsc{IRIS} \citep{li-etal-2025-iris}, the learnable-retrieval method that uses a fixed query set across all patients, as a baseline; this comparison is the exact ablation of patient-conditioned gating, since \textsc{IRIS} shares the same architecture and training setup with all queries activated. Performance metrics are averaged across three runs and shown in Table~\ref{tab:ablation}.

\textsc{EviGen} outperforms the fine-tuned RAG variant on all three datasets, indicating that retrieving by predictive value is superior to retrieving by textual relevance. \textsc{EviGen} also outperforms \textsc{IRIS} on all three datasets, demonstrating the value of patient-conditioned query activation. The gap is narrow on mortality, but widens to 2.7-3.5pp for the real-EHR autism and ADHD datasets that contain longer, sparser inputs. The 32K-context QLoRA variant beats \textsc{EviGen} on mortality, but falls short on autism and ADHD. Moreover, \textsc{EviGen}'s compute cost does not scale with input length, as it operates on a fixed-size retrieved set, whereas full-context QLoRA scales quadratically. All \textsc{EviGen} training runs complete within 30 minutes on an A5000 Ada GPU, compared with 35 to 80 GPU-hours for 32K QLoRA on H200 GPUs.

Beyond these methods, our literature review identified few other supervised approaches directly applicable to outcome prediction over arbitrarily long clinical document sequences, with \textsc{IRIS} remaining the closest and strongest prior baseline for our setting. We nevertheless evaluate three additional supervised baselines: REMed-Chunk \citep{pmlr-v252-kim24a}, ABMIL \citep{ilse2018attention}, and ModernBERT \citep{warner2025smarter} (details in Appendix~\ref{app:supervised}). \textsc{EviGen} significantly outperforms all three across the three tasks, with the bootstrap 95\% confidence interval of the paired AUC difference entirely above zero.

Finally, while our RAG baseline uses clinically informed, expert-reviewed multi-factor queries over both notes and codes, the same query set is fixed across patients, which limits its adaptability. We therefore also evaluate ReAct-RAG, an adaptive retrieval baseline from the EHR-RAG work \citep{cao2026ehr}, in which the LLM iteratively generates new patient-specific queries based on the evidence retrieved so far. \textsc{EviGen} substantially outperforms ReAct-RAG on both prediction and faithfulness on the mortality task (Appendix~\ref{app:supervised}).

\begin{table}[t]
\centering
\footnotesize
\setlength{\tabcolsep}{4pt}
\begin{tabular}{l r r r}
\toprule
\textbf{Method} & Mortality & Autism & ADHD \\
\midrule
\textsc{IRIS}                       & 0.868 & 0.703 & 0.669 \\
QLoRA-8B (32K)            & \textbf{0.881} & 0.663 & 0.560 \\
QLoRA-8B (RAG)            & 0.864 & 0.689 & 0.647 \\
\midrule
\textbf{\textsc{EviGen} (ours)}     & 0.871 & \textbf{0.730} & \textbf{0.704} \\
\bottomrule
\end{tabular}
\caption{Comparison with supervised baselines trained on the same splits and outcome labels. We report AUC on the three datasets, averaged over three runs.}
\label{tab:ablation}
\end{table}

\section{Component Analysis}
\label{sec:component}

Within its three-layer architecture, \textsc{EviGen} makes three key design choices beyond a standard retrieve-then-generate pipeline: (i) \emph{patient-conditioned query activation}, which decides which learned queries fire for each patient; (ii) \emph{predictive evidence selection with attribution ranking}, which selects and orders evidence by contribution to the prediction rather than textual relevance; and (iii) \emph{evidence-scaffolded generation}, which constrains the LLM to quote and reason over the ranked evidence pack. End-to-end results (\S\ref{sec:results}) show that the full system outperforms baselines, but not how much each choice contributes. We isolate them as follows: (i) is isolated by the \textsc{IRIS} comparison in \S\ref{sec:ablation}, its exact ablation; \S\ref{sec:component:2x2} separates the contributions of (ii) and (iii) to rationale faithfulness; and \S\ref{sec:component:ranking} tests (ii) directly, asking whether attribution-ranked evidence is genuinely more predictive.

\subsection{Isolating evidence selection and scaffold}
\label{sec:component:2x2}

\textsc{EviGen}'s faithfulness gains over RAG (\S\ref{sec:results:faithfulness}) could come from two mechanisms: the LLM may receive \emph{better evidence} (ii), or a \emph{better output format} (iii). We disentangle the two with a matched $2\times2$ experiment on the MIMIC-IV mortality test set across four LLM backbones, generating each rationale under one of four conditions defined by two independent choices. The first is \emph{which evidence the LLM sees}: the top-5 chunks retrieved by similarity to the fixed RAG queries, or the top-5 chunks selected and ranked by \textsc{EviGen}'s Integrated Gradients (IG) attribution. The second is \emph{how the LLM writes}: free-form generation over the concatenated chunks, or \textsc{EviGen}'s evidence scaffold, which pairs each reasoning step with a ranked chunk and its attribution score. The two extreme conditions correspond to the RAG baseline (similarity evidence, free-form) and to \textsc{EviGen} (attribution evidence, scaffold); the two mixed conditions expose each factor's individual effect. Across all conditions, we hold fixed the backbone, number of chunks, token budget, output format, decoding setup, and per-patient predicted probability. Full results are in Appendix~\ref{app:component-full}.

Both choices matter: adding the evidence scaffold raises the joint S1$\And$S2 pass rate by 9.7--21.2pp, and switching from similarity-ranked to IG-ranked evidence adds a further 3.4--11.6pp across the four backbones. The gains come mostly from S1 source grounding (up to +23.8pp from the scaffold and +11.2pp from the evidence source), with smaller but consistent gains in S2. We hypothesize that IG-ranked chunks more often contain self-contained clinical facts directly tied to the prediction, making them easier to quote and reason over, whereas similarity-based RAG may retrieve only broadly relevant chunks.

\subsection{Predictive value of evidence rankings}
\label{sec:component:ranking}

We then test the attribution ranking itself: does it place more predictive evidence at the top? For each patient, we construct two matched five-chunk sets, ranked either by IG attribution or by cosine similarity to the fixed RAG queries, and require the \textsc{EviGen} predictor and each LLM backbone to predict from one set alone (Table~\ref{tab:ranking-value}, Appendix~\ref{app:component-full}). IG-ranked evidence yields substantially higher AUC: 0.868 versus 0.661 for the \textsc{EviGen} predictor, and gains of 13.8--21.4pp for the four LLM backbones. Notably, although the IG ranking is derived from \textsc{EviGen}'s own predictor, it also substantially benefits the zero-shot LLMs; the top-ranked chunks therefore carry genuinely predictive, transferable information rather than predictor-specific artifacts.

\section{Conclusion and Future Work}
\label{sec:conclusion}
In this paper, we presented \textsc{EviGen}, a three-layer framework that produces verifiable clinical rationales from longitudinal EHRs via predictive evidence retrieval, citation-grounded rationale generation, and process-supervised verification. \textsc{EviGen} outperforms full-context and RAG-based LLM baselines on prediction accuracy, rationale faithfulness, and clinical reviewer preference across three medical tasks. Future work can extend \textsc{EviGen} to additional EHR modalities (e.g., lab values and clinical imaging), grounding clinical rationale generation in a richer perspective of the patient record. We also see room to integrate the verifier into the generation loop itself rather than as a post-hoc check.

\section*{Limitations}

Our clinical reviewer pilot is small-scale, consisting of 7 medical students. This serves as preliminary evidence, but is not statistically powered, and the reviewers are pre-licensure trainees rather than practicing clinicians. A larger study with licensed clinicians is needed to further confirm \textsc{EviGen}'s clinical utility. Such a study takes considerable time due to the administrative process, which is currently underway.

Another limitation is that \textsc{EviGen} requires labeled outcomes to train the query vectors that select predictive evidence. For most tasks, labels can be extracted from structured fields in the EHR. For rare-disease prediction, however, the small number of positive cases may make it difficult to learn useful query vectors, reducing \textsc{EviGen}'s advantage over zero-shot baselines.

Finally, the institutional autism and ADHD datasets contain protected longitudinal clinical notes and cannot be shared, an inherent constraint of research on real longitudinal EHRs. The MIMIC-IV setting is fully reproducible: we release the preprocessing pipeline, training and evaluation scripts, and prompts. While label leakage is controlled by the age cutoffs and cohort construction (Appendix~\ref{app:datasets:peds}), we did not exhaustively audit target-related free-text mentions before the cutoff.

\section*{Ethical Considerations}

\textsc{EviGen} is intended as a decision-support and rationale-auditing tool for clinician review, not an autonomous diagnostic system. Early pediatric ASD/ADHD prediction carries risks in both error directions: false positives can cause unnecessary family distress, stigma, and costs, while false negatives can delay needed evaluation and support. Because the models learn from EHRs, they inherit documentation and access biases; children with sparser records, often correlated with access to care, may be disproportionately missed, so deployment should monitor performance across subgroups rather than rely on aggregate accuracy. Finally, citation-grounded rationales improve auditability but can induce automation bias, where a reviewer over-trusts a fluent, well-cited, yet incorrect rationale. The step-level verifier is meant to counter this by flagging unreliable steps, but it is imperfect and is not a correctness guarantee. We therefore treat prospective validation, subgroup monitoring, and human-in-the-loop use as prerequisites for any clinical deployment.
\section*{Acknowledgments}

This work was supported by the National Institute of Mental Health (K01MH127309; PI Matthew Engelhard). AI assistants were used for coding assistance, editing, and proofreading during manuscript preparation.

\bibliography{anthology,custom}

\begin{thebibliography}{45}
\providecommand{\natexlab}[1]{#1}

\bibitem[{Asgari et~al.(2025)Asgari, Monta{\~n}a-Brown, Dubois, Khalil,
  Balloch, Yeung, and Pimenta}]{asgari2025creola}
Elham Asgari, Nina Monta{\~n}a-Brown, Magda Dubois, Saleh Khalil, Jasmine
  Balloch, Joshua~Au Yeung, and Dominic Pimenta. 2025.
\newblock \href {https://doi.org/10.1038/s41746-025-01670-7} {A framework to
  assess clinical safety and hallucination rates of {LLMs} for medical text
  summarisation}.
\newblock \emph{npj Digital Medicine}, 8(1):274.

\bibitem[{Bengio et~al.(2013)Bengio, L{\'e}onard, and
  Courville}]{bengio2013estimating}
Yoshua Bengio, Nicholas L{\'e}onard, and Aaron Courville. 2013.
\newblock Estimating or propagating gradients through stochastic neurons for
  conditional computation.
\newblock \emph{arXiv preprint arXiv:1308.3432}.

\bibitem[{Brooke et~al.(1996)}]{brooke1996sus}
John Brooke et~al. 1996.
\newblock Sus-a quick and dirty usability scale.
\newblock \emph{Usability evaluation in industry}, 189(194):4--7.

\bibitem[{Cao et~al.(2026)Cao, Chen, and Guo}]{cao2026ehr}
Lang Cao, Qingyu Chen, and Yue Guo. 2026.
\newblock Ehr-rag: Bridging long-horizon structured electronic health records
  and large language models via enhanced retrieval-augmented generation.
\newblock \emph{arXiv preprint arXiv:2601.21340}.

\bibitem[{Chiang and Lee(2023)}]{chiang2023can}
Cheng-Han Chiang and Hung-yi Lee. 2023.
\newblock Can large language models be an alternative to human evaluations?
\newblock In \emph{Proceedings of the 61st Annual Meeting of the Association
  for Computational Linguistics (Volume 1: Long Papers)}, pages 15607--15631.

\bibitem[{Dymek et~al.(2021)Dymek, Kim, Melton, Payne, Singh, and
  Hsiao}]{dymek2021building}
Christine Dymek, Bryan Kim, Genevieve~B Melton, Thomas~H Payne, Hardeep Singh,
  and Chun-Ju Hsiao. 2021.
\newblock Building the evidence-base to reduce electronic health
  record--related clinician burden.
\newblock \emph{Journal of the American Medical Informatics Association},
  28(5):1057--1061.

\bibitem[{Engelhard et~al.(2020)Engelhard, Berchuck, Garg, Henao, Olson,
  Rusincovitch, Dawson, and Kollins}]{engelhard2020util}
Matthew~M. Engelhard, Samuel~I. Berchuck, Jyotsna Garg, Ricardo Henao, Andrew
  Olson, Shelley Rusincovitch, Geraldine Dawson, and Scott~H. Kollins. 2020.
\newblock \href {https://doi.org/10.1038/s41598-020-74458-2} {Health system
  utilization before age 1 among children later diagnosed with autism or
  {ADHD}}.
\newblock \emph{Scientific Reports}, 10(1):17677.

\bibitem[{Engelhard et~al.(2023)Engelhard, Henao, Berchuck, Chen, Eichner,
  Herkert, Kollins, Olson, Perrin, Rogers et~al.}]{engelhard2023predictive}
Matthew~M Engelhard, Ricardo Henao, Samuel~I Berchuck, Junya Chen, Brian
  Eichner, Darby Herkert, Scott~H Kollins, Andrew Olson, Eliana~M Perrin,
  Ursula Rogers, et~al. 2023.
\newblock Predictive value of early autism detection models based on electronic
  health record data collected before age 1 year.
\newblock \emph{JAMA network open}, 6(2):e2254303.

\bibitem[{Gao et~al.(2023)Gao, Dligach, Miller, Churpek, and
  Afshar}]{gao-etal-2023-overview}
Yanjun Gao, Dmitriy Dligach, Timothy Miller, Matthew~M Churpek, and Majid
  Afshar. 2023.
\newblock \href {https://doi.org/10.18653/v1/2023.bionlp-1.43} {Overview of the
  problem list summarization ({P}rob{S}um) 2023 shared task on summarizing
  patients' active diagnoses and problems from electronic health record
  progress notes}.
\newblock In \emph{Proceedings of the 22nd Workshop on Biomedical Natural
  Language Processing and BioNLP Shared Tasks}, pages 461--467, Toronto,
  Canada. Association for Computational Linguistics.

\bibitem[{Ghorayeb et~al.(2023)Ghorayeb, Darbyshire, Wronikowska, and
  Watkinson}]{ghorayeb2023design}
Abir Ghorayeb, Julie~L Darbyshire, Marta~W Wronikowska, and Peter~J Watkinson.
  2023.
\newblock Design and validation of a new healthcare systems usability scale
  (hsus) for clinical decision support systems: a mixed-methods approach.
\newblock \emph{BMJ open}, 13(1):e065323.

\bibitem[{Goh et~al.(2024)Goh, Gallo, Hom, Strong, Weng, Kerman, Cool, Kanjee,
  Parsons, Ahuja et~al.}]{goh2024large}
Ethan Goh, Robert Gallo, Jason Hom, Eric Strong, Yingjie Weng, Hannah Kerman,
  Jos{\'e}phine~A Cool, Zahir Kanjee, Andrew~S Parsons, Neera Ahuja, et~al.
  2024.
\newblock Large language model influence on diagnostic reasoning: a randomized
  clinical trial.
\newblock \emph{JAMA network open}, 7(10):e2440969.

\bibitem[{Grattafiori et~al.(2024)Grattafiori, Dubey, Jauhri, Pandey, Kadian,
  Al-Dahle, Letman, Mathur, Schelten, Vaughan et~al.}]{grattafiori2024llama}
Aaron Grattafiori, Abhimanyu Dubey, Abhinav Jauhri, Abhinav Pandey, Abhishek
  Kadian, Ahmad Al-Dahle, Aiesha Letman, Akhil Mathur, Alan Schelten, Alex
  Vaughan, et~al. 2024.
\newblock The llama 3 herd of models.
\newblock \emph{arXiv preprint arXiv:2407.21783}.

\bibitem[{Guo et~al.(2025)Guo, Cheng, Tang, Tu, and Lin}]{guo2025dynamic}
Yongxin Guo, Zhenglin Cheng, Xiaoying Tang, Zhaopeng Tu, and Tao Lin. 2025.
\newblock Dynamic mixture of experts: An auto-tuning approach for efficient
  transformer models.
\newblock In \emph{International Conference on Learning Representations},
  volume 2025, pages 79643--79672.

\bibitem[{Hoffman et~al.(2023)Hoffman, Mueller, Klein, and
  Litman}]{hoffman2023measures}
Robert~R Hoffman, Shane~T Mueller, Gary Klein, and Jordan Litman. 2023.
\newblock Measures for explainable ai: Explanation goodness, user satisfaction,
  mental models, curiosity, trust, and human-ai performance.
\newblock \emph{Frontiers in Computer Science}, 5:1096257.

\bibitem[{Holzinger et~al.(2020)Holzinger, Carrington, and
  M{\"u}ller}]{holzinger2020measuring}
Andreas Holzinger, Andr{\'e} Carrington, and Heimo M{\"u}ller. 2020.
\newblock Measuring the quality of explanations: the system causability scale
  (scs) comparing human and machine explanations.
\newblock \emph{KI-K{\"u}nstliche Intelligenz}, 34(2):193--198.

\bibitem[{Ilse et~al.(2018)Ilse, Tomczak, and Welling}]{ilse2018attention}
Maximilian Ilse, Jakub Tomczak, and Max Welling. 2018.
\newblock Attention-based deep multiple instance learning.
\newblock In \emph{International conference on machine learning}, pages
  2127--2136. Pmlr.

\bibitem[{Johnson et~al.(2023)Johnson, Bulgarelli, Shen, Gayles, Shammout,
  Horng, Pollard, Hao, Moody, Gow et~al.}]{johnson2023mimic}
Alistair~EW Johnson, Lucas Bulgarelli, Lu~Shen, Alvin Gayles, Ayad Shammout,
  Steven Horng, Tom~J Pollard, Sicheng Hao, Benjamin Moody, Brian Gow, et~al.
  2023.
\newblock Mimic-iv, a freely accessible electronic health record dataset.
\newblock \emph{Scientific data}, 10(1):1.

\bibitem[{Kim et~al.(2024)Kim, Shim, Yang, Im, Lim, Jeong, and
  Choi}]{pmlr-v252-kim24a}
Junu Kim, Chaeeun Shim, Bosco Seong~Kyu Yang, Chami Im, Sung~Yoon Lim, Han-Gil
  Jeong, and Edward Choi. 2024.
\newblock \href {https://proceedings.mlr.press/v252/kim24a.html}
  {General-purpose retrieval-enhanced medical prediction model using
  near-infinite history}.
\newblock In \emph{Proceedings of the 9th Machine Learning for Healthcare
  Conference}, volume 252 of \emph{Proceedings of Machine Learning Research}.
  PMLR.

\bibitem[{Kirby et~al.(2016)Kirby, Speltz, Rasmussen, Basford, Gottesman,
  Peissig, Pacheco, Tromp, Pathak, Carrell et~al.}]{kirby2016phekb}
Jacqueline~C Kirby, Peter Speltz, Luke~V Rasmussen, Melissa Basford, Omri
  Gottesman, Peggy~L Peissig, Jennifer~A Pacheco, Gerard Tromp, Jyotishman
  Pathak, David~S Carrell, et~al. 2016.
\newblock Phekb: a catalog and workflow for creating electronic phenotype
  algorithms for transportability.
\newblock \emph{Journal of the American Medical Informatics Association},
  23(6):1046--1052.

\bibitem[{Kruse et~al.(2025)Kruse, Hu, Derby, Wu, Stonbraker, Yao, Wang,
  Goldberg, and Gao}]{kruse-etal-2025-large}
Maya Kruse, Shiyue Hu, Nicholas Derby, Yifu Wu, Samantha Stonbraker, Bingsheng
  Yao, Dakuo Wang, Elizabeth~M. Goldberg, and Yanjun Gao. 2025.
\newblock \href {https://doi.org/10.18653/v1/2025.findings-emnlp.1128} {Large
  language models with temporal reasoning for longitudinal clinical
  summarization and prediction}.
\newblock In \emph{Findings of the Association for Computational Linguistics:
  EMNLP 2025}, pages 20715--20735, Suzhou, China. Association for Computational
  Linguistics.

\bibitem[{Lewis et~al.(2020)Lewis, Perez, Piktus, Petroni, Karpukhin, Goyal,
  K{\"u}ttler, Lewis, Yih, Rockt{\"a}schel et~al.}]{lewis2020retrieval}
Patrick Lewis, Ethan Perez, Aleksandra Piktus, Fabio Petroni, Vladimir
  Karpukhin, Naman Goyal, Heinrich K{\"u}ttler, Mike Lewis, Wen-tau Yih, Tim
  Rockt{\"a}schel, et~al. 2020.
\newblock Retrieval-augmented generation for knowledge-intensive nlp tasks.
\newblock \emph{Advances in neural information processing systems},
  33:9459--9474.

\bibitem[{Li et~al.(2025)Li, Hill, Shu, Gao, and Engelhard}]{li-etal-2025-iris}
Fengnan Li, Elliot~D. Hill, Jiang Shu, Jiaxin Gao, and Matthew~M. Engelhard.
  2025.
\newblock \href {https://doi.org/10.18653/v1/2025.acl-long.1461} {{IRIS}:
  Interpretable retrieval-augmented classification for long interspersed
  document sequences}.
\newblock In \emph{Proceedings of the 63rd Annual Meeting of the Association
  for Computational Linguistics (Volume 1: Long Papers)}, pages 30263--30283,
  Vienna, Austria. Association for Computational Linguistics.

\bibitem[{Li et~al.(2022)Li, Ma, and Gao}]{li2022integrating}
Rui Li, Fenglong Ma, and Jing Gao. 2022.
\newblock Integrating multimodal electronic health records for diagnosis
  prediction.
\newblock In \emph{AMIA Annual Symposium Proceedings}, volume 2021, page 726.

\bibitem[{Lightman et~al.(2024)Lightman, Kosaraju, Burda, Edwards, Baker, Lee,
  Leike, Schulman, Sutskever, and Cobbe}]{lightman2024let}
Hunter Lightman, Vineet Kosaraju, Yuri Burda, Harrison Edwards, Bowen Baker,
  Teddy Lee, Jan Leike, John Schulman, Ilya Sutskever, and Karl Cobbe. 2024.
\newblock Let's verify step by step.
\newblock In \emph{International Conference on Learning Representations},
  volume 2024, pages 39578--39601.

\bibitem[{Liu et~al.(2024)Liu, Lin, Hewitt, Paranjape, Bevilacqua, Petroni, and
  Liang}]{liu-etal-2024-lost}
Nelson~F. Liu, Kevin Lin, John Hewitt, Ashwin Paranjape, Michele Bevilacqua,
  Fabio Petroni, and Percy Liang. 2024.
\newblock \href {https://doi.org/10.1162/tacl_a_00638} {Lost in the middle: How
  language models use long contexts}.
\newblock \emph{Transactions of the Association for Computational Linguistics},
  12:157--173.

\bibitem[{Loh et~al.(2025)Loh, Hill, Liu, Dawson, and
  Engelhard}]{loh2025limitations}
De~Rong Loh, Elliot~D Hill, Nan Liu, Geraldine Dawson, and Matthew~M Engelhard.
  2025.
\newblock Limitations of binary classification for long-horizon diagnosis
  prediction and advantages of a discrete-time time-to-event approach:
  Empirical analysis.
\newblock \emph{JMIR AI}, 4:e62985.

\bibitem[{Lopez et~al.(2025)Lopez, Swaminathan, Vedula, Narayanan,
  Nateghi~Haredasht, Ma, Liang, Tate, Maddali, Gallo
  et~al.}]{lopez2025clinical}
Ivan Lopez, Akshay Swaminathan, Karthik Vedula, Sanjana Narayanan, Fateme
  Nateghi~Haredasht, Stephen~P Ma, April~S Liang, Steven Tate, Manoj Maddali,
  Robert~Joseph Gallo, et~al. 2025.
\newblock Clinical entity augmented retrieval for clinical information
  extraction.
\newblock \emph{NPJ digital medicine}, 8(1):45.

\bibitem[{Ma et~al.(2023)Ma, Gong, He, Zhao, and Duan}]{ma2023query}
Xinbei Ma, Yeyun Gong, Pengcheng He, Hai Zhao, and Nan Duan. 2023.
\newblock Query rewriting in retrieval-augmented large language models.
\newblock In \emph{Proceedings of the 2023 Conference on Empirical Methods in
  Natural Language Processing}, pages 5303--5315.

\bibitem[{Mohsen et~al.(2022)Mohsen, Ali, El~Hajj, and
  Shah}]{mohsen2022artificial}
Farida Mohsen, Hazrat Ali, Nady El~Hajj, and Zubair Shah. 2022.
\newblock Artificial intelligence-based methods for fusion of electronic health
  records and imaging data.
\newblock \emph{Scientific Reports}, 12(1):17981.

\bibitem[{Qin et~al.(2025)Qin, Li, Li, Xu, Shi, Lin, Cui, Li, and
  Sun}]{qin2026incentivizing}
Yulei Qin, Gang Li, Zongyi Li, Zihan Xu, Yuchen Shi, Zhekai Lin, Xiao Cui,
  Ke~Li, and Xing Sun. 2025.
\newblock Incentivizing reasoning for advanced instruction-following of large
  language models.
\newblock \emph{Advances in Neural Information Processing Systems},
  38:108337--108401.

\bibitem[{Rule et~al.(2021)Rule, Bedrick, Chiang, and Hribar}]{rule2021length}
Adam Rule, Steven Bedrick, Michael~F Chiang, and Michelle~R Hribar. 2021.
\newblock Length and redundancy of outpatient progress notes across a decade at
  an academic medical center.
\newblock \emph{JAMA Network Open}, 4(7):e2115334.

\bibitem[{Schick and Sch{\"u}tze(2021)}]{schick-schutze-2021-exploiting}
Timo Schick and Hinrich Sch{\"u}tze. 2021.
\newblock \href {https://doi.org/10.18653/v1/2021.eacl-main.20} {Exploiting
  cloze-questions for few-shot text classification and natural language
  inference}.
\newblock In \emph{Proceedings of the 16th Conference of the European Chapter
  of the Association for Computational Linguistics: Main Volume}, pages
  255--269, Online. Association for Computational Linguistics.

\bibitem[{Sohn et~al.(2025)Sohn, Park, Yoon, Park, Hwang, Sung, Kim, and
  Kang}]{sohn-etal-2025-rationale}
Jiwoong Sohn, Yein Park, Chanwoong Yoon, Sihyeon Park, Hyeon Hwang, Mujeen
  Sung, Hyunjae Kim, and Jaewoo Kang. 2025.
\newblock \href {https://doi.org/10.18653/v1/2025.naacl-long.635}
  {Rationale-guided retrieval augmented generation for medical question
  answering}.
\newblock In \emph{Proceedings of the 2025 Conference of the Nations of the
  Americas Chapter of the Association for Computational Linguistics: Human
  Language Technologies (Volume 1: Long Papers)}, pages 12739--12753,
  Albuquerque, New Mexico. Association for Computational Linguistics.

\bibitem[{Sundararajan et~al.(2017)Sundararajan, Taly, and
  Yan}]{sundararajan2017axiomatic}
Mukund Sundararajan, Ankur Taly, and Qiqi Yan. 2017.
\newblock Axiomatic attribution for deep networks.
\newblock In \emph{International conference on machine learning}, pages
  3319--3328. PMLR.

\bibitem[{Van~Veen et~al.(2024)Van~Veen, Van~Uden, Blankemeier, Delbrouck,
  Aali, Bluethgen, Pareek, Polacin, Reis, Seehofnerov{\'a}
  et~al.}]{van2024adapted}
Dave Van~Veen, Cara Van~Uden, Louis Blankemeier, Jean-Benoit Delbrouck, Asad
  Aali, Christian Bluethgen, Anuj Pareek, Malgorzata Polacin, Eduardo~Pontes
  Reis, Anna Seehofnerov{\'a}, et~al. 2024.
\newblock Adapted large language models can outperform medical experts in
  clinical text summarization.
\newblock \emph{Nature medicine}, 30(4):1134--1142.

\bibitem[{Wang et~al.(2025)Wang, Gao, Xu, Liu, Hussein, Korsapati, El~Labban,
  Iheasirim, Hassan, Anil, Bartlett, and Sun}]{wang-etal-2025-process}
Hanyin Wang, Chufan Gao, Qiping Xu, Bolun Liu, Guleid Hussein, Hariprasad~Reddy
  Korsapati, Mohamad El~Labban, Kingsley Iheasirim, Mohamed Hassan, Gokhan
  Anil, Brian Bartlett, and Jimeng Sun. 2025.
\newblock \href {https://doi.org/10.18653/v1/2025.emnlp-main.967}
  {Process-supervised reward models for verifying clinical note generation: A
  scalable approach guided by domain expertise}.
\newblock In \emph{Proceedings of the 2025 Conference on Empirical Methods in
  Natural Language Processing}, pages 19127--19147, Suzhou, China. Association
  for Computational Linguistics.

\bibitem[{Wang et~al.(2023)Wang, Yang, and Wei}]{wang2023query2doc}
Liang Wang, Nan Yang, and Furu Wei. 2023.
\newblock Query2doc: Query expansion with large language models.
\newblock In \emph{Proceedings of the 2023 Conference on Empirical Methods in
  Natural Language Processing}, pages 9414--9423.

\bibitem[{Warner et~al.(2025)Warner, Chaffin, Clavi{\'e}, Weller,
  Hallstr{\"o}m, Taghadouini, Gallagher, Biswas, Ladhak, Aarsen
  et~al.}]{warner2025smarter}
Benjamin Warner, Antoine Chaffin, Benjamin Clavi{\'e}, Orion Weller, Oskar
  Hallstr{\"o}m, Said Taghadouini, Alexis Gallagher, Raja Biswas, Faisal
  Ladhak, Tom Aarsen, et~al. 2025.
\newblock Smarter, better, faster, longer: A modern bidirectional encoder for
  fast, memory efficient, and long context finetuning and inference.
\newblock In \emph{Proceedings of the 63rd annual meeting of the association
  for computational linguistics (volume 1: Long papers)}, pages 2526--2547.

\bibitem[{Williams et~al.(2024)Williams, Miao, Kornblith, and
  Butte}]{williams2024evaluating}
Christopher~YK Williams, Brenda~Y Miao, Aaron~E Kornblith, and Atul~J Butte.
  2024.
\newblock Evaluating the use of large language models to provide clinical
  recommendations in the emergency department.
\newblock \emph{Nature communications}, 15(1):8236.

\bibitem[{Wu et~al.(2024)Wu, Zhu, Qi, Chen, Xu, Menolascina, and
  Grau}]{wu2024medical}
Junde Wu, Jiayuan Zhu, Yunli Qi, Jingkun Chen, Min Xu, Filippo Menolascina, and
  Vicente Grau. 2024.
\newblock Medical graph rag: Towards safe medical large language model via
  graph retrieval-augmented generation.
\newblock \emph{arXiv preprint arXiv:2408.04187}.

\bibitem[{Xiong et~al.(2024{\natexlab{a}})Xiong, Jin, Lu, and
  Zhang}]{xiong-etal-2024-benchmarking}
Guangzhi Xiong, Qiao Jin, Zhiyong Lu, and Aidong Zhang. 2024{\natexlab{a}}.
\newblock \href {https://doi.org/10.18653/v1/2024.findings-acl.372}
  {Benchmarking retrieval-augmented generation for medicine}.
\newblock In \emph{Findings of the Association for Computational Linguistics:
  ACL 2024}, pages 6233--6251, Bangkok, Thailand. Association for Computational
  Linguistics.

\bibitem[{Xiong et~al.(2024{\natexlab{b}})Xiong, Jin, Wang, Zhang, Lu, and
  Zhang}]{xiong2024improving}
Guangzhi Xiong, Qiao Jin, Xiao Wang, Minjia Zhang, Zhiyong Lu, and Aidong
  Zhang. 2024{\natexlab{b}}.
\newblock Improving retrieval-augmented generation in medicine with iterative
  follow-up questions.
\newblock In \emph{Biocomputing 2025: Proceedings of the Pacific Symposium},
  pages 199--214. World Scientific.

\bibitem[{Yang et~al.(2025)Yang, Li, Yang, Zhang, Hui, Zheng, Yu, Gao, Huang,
  Lv et~al.}]{yang2025qwen3}
An~Yang, Anfeng Li, Baosong Yang, Beichen Zhang, Binyuan Hui, Bo~Zheng, Bowen
  Yu, Chang Gao, Chengen Huang, Chenxu Lv, et~al. 2025.
\newblock Qwen3 technical report.
\newblock \emph{arXiv preprint arXiv:2505.09388}.

\bibitem[{Zhao et~al.(2025)Zhao, Liu, Yang, and Miao}]{zhao2025medrag}
Xuejiao Zhao, Siyan Liu, Su-Yin Yang, and Chunyan Miao. 2025.
\newblock Medrag: Enhancing retrieval-augmented generation with knowledge
  graph-elicited reasoning for healthcare copilot.
\newblock In \emph{Proceedings of the ACM on Web Conference 2025}, pages
  4442--4457.

\bibitem[{Zhu et~al.(2024)Zhu, Ren, Wang, Zheng, Xie, Feng, Zhu, Li, Ma, and
  Pan}]{zhu2024emerge}
Yinghao Zhu, Changyu Ren, Zixiang Wang, Xiaochen Zheng, Shiyun Xie, Junlan
  Feng, Xi~Zhu, Zhoujun Li, Liantao Ma, and Chengwei Pan. 2024.
\newblock Emerge: Enhancing multimodal electronic health records predictive
  modeling with retrieval-augmented generation.
\newblock In \emph{Proceedings of the 33rd ACM International Conference on
  Information and Knowledge Management}, pages 3549--3559.

\end{thebibliography}

\appendix

\clearpage
\onecolumn
\section*{Appendix}
\startcontents[appendix]
\printcontents[appendix]{l}{1}{\setcounter{tocdepth}{3}}
\clearpage
\twocolumn

\section{\textsc{EviGen} Implementation Details}
\label{app:evigen-impl}

\subsection{Evidence Selection Layer}
\label{app:evidence-layer}

\subsubsection{Encoder and input}
\label{app:a1:encoder}

\paragraph{Text encoder.} All chunks (notes and ICD codes) are embedded with Qwen3-Embedding-8B, producing $d = 4096$-dim vectors.

\paragraph{Input formatting.} Chunk size 200 tokens with overlap, each chunk prefixed by an age-aware header (\texttt{"Patient age: X years\textbackslash n\textbackslash nClinical Note:\textbackslash n..."}) so the embedding carries demographic context. ICD codes are embedded using the official description with age prefix.

\subsubsection{Architecture and gating}
\label{app:a1:gating}

\paragraph{Architecture dimensions.} The number of queries depends on the task: $N_n = N_c = 4$ for mortality (so $N = 8$ total) and $N_n = N_c = 6$ for ADHD and autism (so $N = 12$). One note query, $\mathbf{q}_0$, is always-on and shared across all patients. Each per-query expert MLP $f_i$ is a pre-LayerNorm 2-layer feed-forward network with GELU activation and a residual connection,
\[
\mathbf{h}_i = \mathbf{c}_i + f_i(\mathrm{LN}(\mathbf{c}_i)),
\]
with hidden dimension $d_h = 1024$ and dropout $0.1$. A final LayerNorm precedes the classifier head.

\paragraph{Straight-through estimator implementation.}\label{app:a1:ste} Let $s = \sigma(\tilde{\mathbf{s}}^{(m)\top}\tilde{\mathbf{q}}_i) - \sigma(\eta_i)$ denote the continuous gating score and $g = \mathbf{1}[s > 0]$ its hard indicator. We implement the STE with the standard stop-gradient identity $g^{\text{STE}} = (g - s)\,\texttt{.detach()} + s$. In the forward pass the two $s$ terms cancel and $g^{\text{STE}} = g$, so retrieval uses the discrete gate. In the backward pass \texttt{.detach()} blocks gradient through the hard branch, leaving $\partial g^{\text{STE}} / \partial \theta = \partial s / \partial \theta$ for any parameter $\theta$, so gradients reach both the query $\mathbf{q}_i$ and the threshold $\eta_i$.

\paragraph{Average active queries.} On the test dataset, the trained gates activate on average 3.48 of the 8 queries per patient for mortality, 6.64 of the 12 for autism, and 9.03 of the 12 for ADHD.

\paragraph{Query-level activation rates.} Table~\ref{tab:activation-rates} reports how often each query is activated across the mortality test set. Activation varies substantially across queries: some are activated for nearly all patients, whereas others are rarely or never activated. Several code queries in particular have low activation rates, suggesting that the model relies more heavily on note-based evidence for this task, or finds some code queries redundant. Similar query-specific patterns are observed on the autism and ADHD datasets.

\begin{table}[t]
\centering
\footnotesize
\setlength{\tabcolsep}{4pt}
\begin{tabular}{c c c r}
\toprule
\textbf{Query} & \textbf{Modality} & \textbf{Gate type} & \textbf{Activation (\%)} \\
\midrule
0 & Note & Shared, always on & 100.0 \\
1 & Note & Dynamic           & 100.0 \\
2 & Note & Dynamic           & 23.9 \\
3 & Note & Dynamic           & 50.6 \\
4 & Code & Dynamic           & 0.0 \\
5 & Code & Dynamic           & 34.6 \\
6 & Code & Dynamic           & 0.0 \\
7 & Code & Dynamic           & 39.3 \\
\bottomrule
\end{tabular}
\caption{Per-query activation rates on the mortality test set.}
\label{tab:activation-rates}
\end{table}

\subsubsection{Retrieval and query learning}
\label{app:a1:retrieval}

\paragraph{Retrieval.} Each active query retrieves the top-$K$ chunks ($K = 4$) by cosine similarity, deterministically at both training and inference; the retrieval width auto-shrinks if a patient has fewer than $K$ valid chunks. Gating uses the hard 0/1 indicator at forward time.

\paragraph{Query learning dynamic.} Although top-$K$ retrieval by cosine similarity is non-differentiable, each query $\mathbf{q}_i$ still receives gradient signal through the attention weights $\alpha_{i,j}$ over the retrieved chunks.

Intuitively, training proceeds as follows. At initialization, $\mathbf{q}_i$ is random and the top-$K$ chunks it retrieves are a mix of predictive and uninformative ones. The loss gradient then pushes the attention weight $\alpha_{i,j}$ upward on chunks whose embeddings would reduce the loss. Because $\alpha_{i,j}$ is a softmax of $\tilde{\mathbf{q}}_i^\top \mathbf{x}_j^{(m)}$, assigning larger weight to a chunk is equivalent to pulling $\mathbf{q}_i$ toward that chunk's embedding direction. After the update, $\mathbf{q}_i$ has moved toward the predictive region of embedding space, so on the next batch the hard top-$K$ retrieval is more likely to surface predictive chunks in the first place. Iterating this loop, $\mathbf{q}_i$ converges to a stable region whose top-$K$ retrieval consistently yields predictive evidence.

Applying the chain rule through Eq.~\ref{eq:attention} confirms this picture (dropping the $\tilde{\cdot}$ for clarity):
\begin{equation}
\begin{aligned}
\frac{\partial \mathcal{L}}{\partial \mathbf{q}_i} &= \sum_{j=1}^{K} \alpha_{i,j}\, u_{i,j}\, \left(\mathbf{x}_j^{(m)} - \mathbf{c}_i\right), \\
u_{i,j} &:= \left(\frac{\partial \mathcal{L}}{\partial \mathbf{c}_i}\right)^{\!\top} \mathbf{x}_j^{(m)}.
\end{aligned}
\label{eq:query-grad}
\end{equation}
The scalar $u_{i,j}$ measures how much chunk $j$'s embedding aligns with the direction in which $\mathbf{c}_i$ would need to change to reduce the loss: $u_{i,j} < 0$ identifies predictive chunks. Gradient descent then shifts $\mathbf{q}_i$ in the direction of $(\mathbf{x}_j^{(m)} - \mathbf{c}_i)$ for chunks with negative $u_{i,j}$, weighted by the current attention $\alpha_{i,j}$, precisely the ``pulling toward predictive embeddings'' described above. \citet{li-etal-2025-iris} provides a detailed convergence analysis.

\subsubsection{Training configuration}
\label{app:a1:training}

\paragraph{Training objective.} Query vectors, per-query gating thresholds, per-query MLPs, and the prediction head are trained end-to-end with cross-entropy loss $\mathcal{L}_{\text{CE}}$ on clinical outcome labels, augmented by an auxiliary loss:
\begin{equation}
\mathcal{L}_{\text{aux}} = \underbrace{\|\tilde{\mathbf{Q}}\tilde{\mathbf{Q}}^\top - \mathbf{I}\|_F^2}_{\text{diversity}} + \underbrace{\tfrac{1}{N-1}\textstyle\sum_{i \geq 1}\|\mathbf{q}_i\|^2}_{\text{simplicity}}
\label{eq:training-loss}
\end{equation}
Here $\tilde{\mathbf{Q}} \in \mathbb{R}^{N \times d}$ stacks the L2-normalized query vectors $\tilde{\mathbf{q}}_i$ as rows, so $\tilde{\mathbf{Q}}\tilde{\mathbf{Q}}^\top$ is the matrix of pairwise cosine similarities between queries; $\mathbf{I}$ is the identity matrix; $\|\cdot\|_F$ denotes the Frobenius norm; and the simplicity-term sum $\sum_{i \geq 1}$ runs over the non-shared queries, excluding the always-on query $\mathbf{q}_0$. The diversity term pushes the Gram matrix toward the identity, preventing different queries from collapsing onto the same pattern; the simplicity term bounds query norms for numerical stability. This formulation adapts the regularizer used for expert representation matrices in dynamic mixture-of-experts \citep{guo2025dynamic}.

\paragraph{Hyperparameters.} We use auxiliary loss coefficient $\lambda_{\text{aux}} = 0.01$ and train with AdamW for up to $E = 20$ epochs, with a task-specific learning rate ($10^{-3}$ for mortality, $3\times 10^{-4}$ for autism, and $10^{-4}$ for ADHD) and early stopping on validation AUC (patience 3); we select the best checkpoint by maximum validation AUC.

\subsection{Rationale Generation Layer}
\label{app:generation-layer}

\paragraph{LLM prompt template.} Table~\ref{tab:prompt} summarizes the structure of the prompt used by the rationale-generation LLM (mortality task variant); the ADHD and autism-prediction variants substitute task-specific framing for the \textit{Task} row. Full verbatim prompt strings are available in the code repository.

\subsection{Process Verification Layer}
\label{app:verification-layer}

\paragraph{Verifier performance.}
We achieve a step-level AUROC of 0.978 for binary error detection, with a calibrated F1 of 0.947 at threshold 0.5 and an ECE of 0.028 after Platt scaling. When flagged steps are typed via conditional renormalization over the five clinical error classes, the verifier assigns the correct error type 88.2\% of the time. At the sample level, the product of calibrated step scores ranks gold rationales above their paired negatives in 98.8\% of cases.

\paragraph{Step-delimiter scheme.}
We repurpose five of Llama-3.1's 256 reserved special tokens as structural delimiters: a \emph{step} trigger after each reasoning step, a \emph{completeness} trigger after each rationale section, an \emph{end-of-note} trigger at the rationale end, a \emph{context-end} boundary between patient evidence and rationale, and a \emph{section} marker before each rationale header.
Each trigger is followed by a label token drawn from a disjoint set of seven reserved tokens: \emph{correct}, \emph{hallucination}, \emph{factual\_inaccuracy}, \emph{unhelpfulness}, \emph{generic\_negative}, \emph{attribution\_distortion}, and \emph{provenance\_break}.
At inference, the model's softmax over these seven token IDs at each trigger position produces a per-step label distribution.

\paragraph{Synthetic training data.}
Gold rationales from \S\ref{sec:method:generation} provide positive examples (all steps labeled \emph{correct}).
For each gold rationale, we generate 30 negative variants by prompting GPT-4o-mini to corrupt 3-7 randomly selected steps, each according to one of five error types:
\begin{itemize}[nosep,leftmargin=1.5em]
  \item \textbf{Hallucination}: inject a clinical claim absent from the patient evidence.
  \item \textbf{Factual inaccuracy}: alter a numeric value, date, or clinical detail present in the evidence.
  \item \textbf{Unhelpfulness}: replace a specific, evidence-supported claim with a vague or speculative statement.
  \item \textbf{Attribution distortion}: change the sign or magnitude of an attribution score.
  \item \textbf{Provenance break}: reassign a citation to a different source document.
\end{itemize}
Corrupted steps receive the matching error-type label; uncorrupted steps retain \emph{correct}.
Clean steps are additionally paraphrased with probability 0.4 to prevent the model from relying on surface-level template matching.
With probability 0.25, 1-2 uncorrupted steps are deleted from the negative sample to train the completeness trigger to detect missing content.
Section-level \emph{completeness} and note-level \emph{end-of-note} triggers are labeled \emph{generic\_negative} if any step in their scope is corrupted or deleted, and \emph{correct} otherwise.
The final training set comprises 1{,}800 patients $\times$ 3 gold rationales $\times$ (1 + 30 negatives) $=$ 167{,}400 samples.

\paragraph{QLoRA configuration.}
We fine-tune Llama-3.1-8B-Instruct with QLoRA: rank $r\!=\!16$, scaling $\alpha\!=\!32$, dropout $0.05$, 4-bit NF4 quantization with double quantization, targeting the \texttt{q\_proj}, \texttt{k\_proj}, \texttt{v\_proj}, and \texttt{o\_proj} attention matrices. The embedding and LM head layers are trained in full precision alongside the LoRA adapters.
The loss function is \texttt{notes\_only}: all tokens after the context-end delimiter contribute to the causal language-modeling objective, and the patient-context prefix is masked.
We train for up to 2 epochs with early stopping; the best checkpoint is consistently at epoch 1. We use 8-bit paged AdamW with learning rate $1.2 \times 10^{-4}$ (effective batch size 64 across 4 GPUs), cosine decay, 100 warmup steps, and max sequence length 2{,}048.

\paragraph{Platt calibration.}
We fit Platt scaling, $\hat{p} = \sigma(a \cdot \mathrm{logit}(p) + b)$, on a 200-patient held-out test set. Patient-level 5-fold cross-validation confirms parameter stability ($a = 1.648 \pm 0.016$, $b = 10.41 \pm 0.09$); final parameters are fit on all held-out data.
After calibration, the correct-step median rises to 0.963 and the corrupted-step median to 0.375, with ECE $= 0.028$ (from 0.780) and step-level F1 $= 0.947$ at threshold 0.5.

\paragraph{Error typing.}
Raw 7-class argmax accuracy is 63.7\%, dominated by the \emph{generic\_negative} prior (${\sim}$51\% of probability mass).
We use a two-stage approach: (1)~flag steps where calibrated $P(\text{correct}) < 0.5$, then (2)~type the error by dropping \emph{correct} and \emph{generic\_negative} from the softmax and renormalizing over the five clinical error types.
Conditional typing accuracy on flagged steps is 88.2\%.

\paragraph{Scoring aggregation.}
The sample-level verification score is the product of calibrated $P(\text{correct})$ across content steps only (completeness and end-of-note triggers are excluded).
Post-calibration pairwise preference accuracy (fraction of gold-negative pairs from the same patient where the gold scores higher) is 98.8\%.

\paragraph{Evaluation on naturally generated rationales.}
Synthetic corruptions alone do not fully establish how the verifier performs on naturally generated outputs. Because obtaining reliable step-level labels requires expert human annotation, we use a judge-based ranking evaluation on naturally generated rationales as a practical alternative. We sampled 300 MIMIC-IV mortality patients and used Llama-3.1-70B to generate 10 rationale rollouts per patient at temperature 1. The verifier scored each rollout by multiplying $P(\text{correct})$ across all reasoning-step delimiters. We then ranked the 10 rollouts for each patient from highest to lowest verifier score and evaluated them using the same local reasoning-consistency judge as in our faithfulness evaluation (Appendix~\ref{app:faithfulness:stage2}).

As shown in Table~\ref{tab:verifier-natural}, the lowest-ranked rollouts had a 42.33\% pass rate, compared with 54.14\% on average for ranks 1--9, a difference of $-11.81$ percentage points (95\% CI: $[-17.56, -5.67]$). Although the rank-wise trend is not monotonic, the lowest-scored rollouts have a substantially lower faithfulness pass rate. This provides preliminary evidence that the verifier can identify less faithful naturally generated rationales beyond the synthetic corruptions used for training.

We emphasize that this is a judge-based ranking evaluation rather than a human-annotated step-level or error-type evaluation. We have not conducted a controlled reader study testing whether displaying verifier flags improves human review quality or efficiency, and therefore do not make that claim; the verifier should be regarded as a preliminary screening tool.

\begin{table}[t]
\centering
\footnotesize
\setlength{\tabcolsep}{6pt}
\begin{tabular}{c r}
\toprule
\textbf{Verifier rank} & \textbf{Faithfulness rate (\%)} \\
\midrule
1  & 56.67 \\
2  & 60.33 \\
3  & 56.67 \\
4  & 50.67 \\
5  & 52.00 \\
6  & 54.67 \\
7  & 54.33 \\
8  & 53.00 \\
9  & 49.00 \\
10 & 42.33 \\
\bottomrule
\end{tabular}
\caption{Judge-based faithfulness pass rate of naturally generated rationales, grouped by verifier rank (1 = highest verifier score; 10 rollouts per patient, 300 MIMIC-IV mortality patients, Llama-3.1-70B at temperature 1).}
\label{tab:verifier-natural}
\end{table}

\section{Baseline Implementation Details}
\label{app:baselines}

\paragraph{Serving and decoding.} We serve open-weight models \citep{grattafiori2024llama, yang2025qwen3} via vLLM with the per-model configuration in Table~\ref{tab:baseline-config}; GPT-4o-mini is accessed via the Azure OpenAI API. When the prompt (system message, user instruction, and patient history) exceeds the input budget, we left-truncate the patient history, keeping the most recent notes and codes. Truncation uses each model's native tokenizer. All open-weight and API-based language models were used in accordance with their respective licenses and terms of service.

\begin{table*}[t]
\small
\centering
\begin{tabular}{l r r}
\toprule
\textbf{Model} & \textbf{input budget} & \textbf{max-new-tokens} \\
\midrule
Llama-3.1-8B-Instruct  & 122{,}880 & 4{,}096  \\
Llama-3.1-70B-Instruct & 122{,}880 & 4{,}096  \\
Qwen3-32B (thinking)   & 114{,}688 & 16{,}384 \\
GPT-4o-mini            & 123{,}904 & 4{,}096  \\
\bottomrule
\end{tabular}
\caption{Per-model prompt and generation budgets for baselines. Qwen3-32B reserves a larger output budget to accommodate thinking traces in addition to the JSON rationale.}
\label{tab:baseline-config}
\end{table*}

\paragraph{YaRN context extension.} Qwen3-32B's native context window is 40{,}960 tokens. We extend it to 131{,}072 via YaRN rope-scaling with factor 4 (\texttt{rope\_type=yarn}, \texttt{factor=4.0}, \texttt{original\_max\_position\_embeddings=40960}). We set the vLLM environment variable \texttt{VLLM\_ALLOW\_LONG\_MAX\_MODEL\_LEN=1} to permit a max-model-len above the model's reported \texttt{max\_position\_embeddings}.

\paragraph{Query encoder and chunking.} The RAG baseline reuses \textsc{EviGen}'s chunking and embedding setup (Appendix~\ref{app:a1:encoder}): 200-token chunks with overlap, embedded by Qwen3-Embedding-8B.

\paragraph{Multi-factor RAG queries.} For each task, Claude Opus 4.7 compiles candidate risk factors from the clinical literature; the factors are reviewed by clinical experts and finalized into a fixed list of note queries and code queries. Each query independently retrieves the top-$K = 4$ chunks by cosine similarity, and the retrieved chunks across all queries are concatenated as the RAG context. The per-task queries are listed in Tables~\ref{tab:rag-mortality}, \ref{tab:rag-adhd}, and~\ref{tab:rag-autism}.

\begin{table*}[t]
\small
\centering
\begin{tabular}{c >{\raggedright\arraybackslash}p{0.44\textwidth} >{\raggedright\arraybackslash}p{0.44\textwidth}}
\toprule
\# & \textbf{Note query} & \textbf{Code query} \\
\midrule
1 & End-of-life / palliative / hospice / DNR / DNI / advance directive / goals of care & Cardiac arrest, AMI, stroke, PE, ARDS, aortic dissection, sudden cardiac death \\
\addlinespace
2 & Organ failure / ICU / mechanical ventilation / vasopressors / cardiac arrest / dialysis / MOSF & ESRD, chronic HF, hepatic failure, chronic respiratory failure, dialysis, organ transplant \\
\addlinespace
3 & Multiple chronic conditions / DM with complications / HF / CKD / COPD / cirrhosis / end-stage disease & Malignant neoplasm, metastatic disease, advanced cancer, oncology, carcinoma \\
\addlinespace
4 & Functional decline / debility / failure to thrive / weight loss / malnutrition / bedbound / cognitive decline / dementia & Sepsis, severe sepsis, septic shock, bacteremia, infectious complication \\
\bottomrule
\end{tabular}
\caption{Multi-factor RAG queries for the 1-year mortality task (MIMIC-IV). 4 note queries + 4 code queries; each retrieves 4 chunks, for 32 retrieved chunks per patient.}
\label{tab:rag-mortality}
\end{table*}

\begin{table*}[t]
\small
\centering
\begin{tabular}{c >{\raggedright\arraybackslash}p{0.44\textwidth} >{\raggedright\arraybackslash}p{0.44\textwidth}}
\toprule
\# & \textbf{Note query} & \textbf{Code query} \\
\midrule
1 & Attention difficulties / distractibility / careless mistakes / unable to sustain attention / forgetful / loses things & ADHD subtypes (inattentive, hyperactive-impulsive, combined, other, unspecified) \\
\addlinespace
2 & Hyperactivity / impulsivity / fidgeting / cannot stay seated / blurts out / interrupts / restless & Disruptive behavior disorders (oppositional defiant, intermittent explosive, conduct disorder) \\
\addlinespace
3 & School / academic concerns / learning difficulties / academic underperformance / school refusal & Specific learning disorders (reading, mathematics, written expression) \\
\addlinespace
4 & Behavioral and conduct issues / oppositional / defiant / aggressive / disciplinary problems & Anxiety and mood disorders (generalized anxiety, depression, separation anxiety, social anxiety) \\
\addlinespace
5 & Sleep / routine disruption / difficulty falling asleep / fatigue / restless sleep & Sleep disorders (insomnia, parasomnia, restless-legs, irregular sleep-wake) \\
\addlinespace
6 & Comorbid emotional and developmental concerns / low self-esteem / peer-relationship issues & Other neurodevelopmental disorders (tic, motor, speech, developmental coordination) \\
\bottomrule
\end{tabular}
\caption{Multi-factor RAG queries for the ADHD task. 6 note queries + 6 code queries; each retrieves 4 chunks, for 48 retrieved chunks per patient.}
\label{tab:rag-adhd}
\end{table*}

\begin{table*}[t]
\small
\centering
\begin{tabular}{c >{\raggedright\arraybackslash}p{0.44\textwidth} >{\raggedright\arraybackslash}p{0.44\textwidth}}
\toprule
\# & \textbf{Note query} & \textbf{Code query} \\
\midrule
1 & Speech / language delay / delayed speech / language regression / expressive impairment & Autism spectrum disorder, autistic disorder, Asperger's, pervasive developmental disorder \\
\addlinespace
2 & Repetitive behaviors / restricted interests / sensory sensitivities / hand-flapping / lining up objects & Developmental delay / global developmental delay / mixed developmental disorder \\
\addlinespace
3 & Social communication / eye contact / joint attention / solitary play / social-emotional reciprocity & Intellectual disability (mild / moderate / severe), cognitive impairment \\
\addlinespace
4 & Cognitive / learning concerns / developmental milestones / regression of skills & Comorbid psychiatric (ADHD, anxiety, mood, oppositional defiant) \\
\addlinespace
5 & Pediatric medical / GI symptoms / motor delays / hypotonia / atypical head circumference / feeding issues & Neurological disorders (epilepsy, seizure, cerebral palsy, microcephaly, macrocephaly) \\
\addlinespace
6 & Sleep / behavioral regulation / anxiety / behavior dysregulation / mood problems & Sleep / GI / motor / sensory-processing disorders \\
\bottomrule
\end{tabular}
\caption{Multi-factor RAG queries for the autism task. 6 note queries + 6 code queries; each retrieves 4 chunks, for 48 retrieved chunks per patient.}
\label{tab:rag-autism}
\end{table*}

\paragraph{Full-context zero-shot prompt.} Table~\ref{tab:zeroshot-prompt} gives the verbatim system prompt and user-instruction template for the full-context zero-shot baseline (mortality task variant); the ADHD and autism variants substitute task-specific framing in the same slots.

\paragraph{RAG prompt.} The RAG baseline reuses the zero-shot prompt schema (Table~\ref{tab:zeroshot-prompt}) with three substitutions: ``complete clinical history (\dots in chronological order)'' becomes ``retrieved clinical evidence (\dots most relevant to mortality risk)''; ``patient history'' becomes ``retrieved evidence''; and the requirement to include the note's documented Date and Age in each supporting-evidence line is omitted, since retrieved chunks may not carry intact note headers. All other fields (hard rules, citation format, the per-factor structure, [Reasoning Lens], [Recommendations], and output JSON schema) are identical.

\paragraph{Verbalizer.} The verbalizer baseline reuses the zero-shot prompt schema (Table~\ref{tab:zeroshot-prompt}) with three modifications. The task is rephrased as binary Yes/No classification: the user instruction asks ``Will this patient die within one year after their last discharge?'' and the output format becomes \texttt{Answer: <Yes or No>} followed by \texttt{\{"report": \dots\}} on the next line. The [Prediction + Explanation] section is reworded to state ``the patient is predicted to die within 1 year'' or ``predicted to survive at least 1 year'' instead of a numeric probability, and an added hard rule forbids stating probabilities anywhere in the rationale; all other fields (per-factor structure, citation rule, [Reasoning Lens], [Recommendations]) are unchanged. To recover a probability, we append \texttt{"Answer: "} as an assistant prefill after \texttt{apply\_chat\_template(...)}, so that the model's first generated token is exactly the verbalizer label; we then read the logits at this position and compute $P(\text{positive}) = \exp(z_\text{Yes}) / (\exp(z_\text{Yes}) + \exp(z_\text{No}))$. The default verbalizer pair is ``Yes''/``No'' (tokenized with a leading space in all baseline tokenizers); ``$+$''/``$-$'' and reserved special tokens are supported as alternative verbalizers for ablation.

\section{Additional Baseline Comparisons}
\label{app:supervised}

\subsection{Additional supervised baselines}
\label{app:supervised:sup}

To broaden the supervised comparison in \S\ref{sec:ablation} beyond \textsc{IRIS}, we add \emph{REMed-Chunk}, an adaptation of REMed \citep{pmlr-v252-kim24a} to our note-and-code setting. The original REMed ranks structured EHR events with outcome supervision and predicts from the top-$k$ events; we adapt it by treating each clinical-note chunk and ICD description as an event, using the same input embeddings as \textsc{EviGen}. We also add \emph{ABMIL} \citep{ilse2018attention}, which uses attention pooling over all note and code embeddings to form a patient representation, and \emph{ModernBERT} \citep{warner2025smarter}, an 8K-context supervised encoder applied to the truncated patient record. All three are trained on the same train and validation splits and outcome labels as \textsc{EviGen}.

Table~\ref{tab:supervised-ci} reports AUC for all supervised methods with bootstrap 95\% confidence intervals. An asterisk marks baselines that \textsc{EviGen} significantly outperforms, i.e., the bootstrap 95\% confidence interval of the paired AUC difference lies entirely above zero. Across all three tasks, \textsc{EviGen} outperforms the newly added supervised baselines, with the largest gains on the longer and sparser autism and ADHD cohorts.

\begin{table}[t]
\centering
\footnotesize
\setlength{\tabcolsep}{3pt}
\begin{tabular}{l r r r}
\toprule
\textbf{Method} & Mortality & Autism & ADHD \\
\midrule
\textsc{IRIS}
  & 0.868 & 0.703\rlap{$^*$} & 0.669\rlap{$^*$} \\[-2pt]
  & \scriptsize [0.858, 0.879] & \scriptsize [0.681, 0.726] & \scriptsize [0.645, 0.693] \\
\addlinespace[2pt]
QLoRA-8B (32K)
  & \textbf{0.881} & 0.663\rlap{$^*$} & 0.560\rlap{$^*$} \\[-2pt]
  & \scriptsize [0.871, 0.891] & \scriptsize [0.636, 0.691] & \scriptsize [0.533, 0.586] \\
\addlinespace[2pt]
QLoRA-8B (RAG)
  & 0.864 & 0.689\rlap{$^*$} & 0.647\rlap{$^*$} \\[-2pt]
  & \scriptsize [0.852, 0.875] & \scriptsize [0.664, 0.714] & \scriptsize [0.622, 0.672] \\
\addlinespace[2pt]
REMed-Chunk
  & 0.862\rlap{$^*$} & 0.686\rlap{$^*$} & 0.660\rlap{$^*$} \\[-2pt]
  & \scriptsize [0.850, 0.873] & \scriptsize [0.663, 0.709] & \scriptsize [0.634, 0.685] \\
\addlinespace[2pt]
ABMIL
  & 0.865\rlap{$^*$} & 0.668\rlap{$^*$} & 0.649\rlap{$^*$} \\[-2pt]
  & \scriptsize [0.853, 0.876] & \scriptsize [0.641, 0.695] & \scriptsize [0.623, 0.675] \\
\addlinespace[2pt]
ModernBERT
  & 0.859\rlap{$^*$} & 0.624\rlap{$^*$} & 0.552\rlap{$^*$} \\[-2pt]
  & \scriptsize [0.849, 0.870] & \scriptsize [0.598, 0.650] & \scriptsize [0.524, 0.580] \\
\midrule
\textbf{\textsc{EviGen} (ours)}
  & 0.871 & \textbf{0.730} & \textbf{0.704} \\[-2pt]
  & \scriptsize [0.860, 0.881] & \scriptsize [0.705, 0.754] & \scriptsize [0.680, 0.728] \\
\bottomrule
\end{tabular}
\caption{Full supervised-baseline comparison. We report AUC with bootstrap 95\% confidence intervals on the three datasets. $^*$: \textsc{EviGen} significantly outperforms the baseline (bootstrap 95\% CI of the paired AUC difference entirely above zero). Best per column in \textbf{bold}.}
\label{tab:supervised-ci}
\end{table}

\subsection{ReAct-RAG}
\label{app:supervised:react}

Most existing medical RAG systems retrieve from an external corpus, e.g., MedRAG \citep{zhao2025medrag} and MIRAGE \citep{xiong-etal-2024-benchmarking}, or from an external knowledge graph, e.g., EMERGE \citep{zhu2024emerge}, which is a different task from \textsc{EviGen}'s within-record predictive retrieval. We therefore compare against ReAct-RAG \citep{cao2026ehr}, which interleaves reasoning and retrieval actions: after examining the evidence retrieved so far, the LLM generates a new patient-specific query to seek additional missing evidence, retrieves new chunks, and repeats this process over multiple rounds. Thus, unlike our fixed-query RAG baseline (\S\ref{sec:exp:baselines}), its retrieval queries depend on both the individual patient record and the evidence obtained in previous iterations. We use three retrieval rounds, following the original setup. The full EHR-RAG framework is not directly applicable without substantial adaptation because it is designed for structured event sequences and numeric temporal trajectories rather than free-text note chunks and ICD descriptions.

Tables~\ref{tab:react-pred} and~\ref{tab:react-faith} compare ReAct-RAG with \textsc{EviGen} on the mortality task. \textsc{EviGen} outperforms ReAct-RAG on prediction AUC for every backbone and achieves substantially higher joint faithfulness pass rates.

\begin{table}[t]
\centering
\footnotesize
\setlength{\tabcolsep}{4pt}
\begin{tabular}{l l r}
\toprule
\textbf{Method} & \textbf{Backbone} & \textbf{Mortality AUC} \\
\midrule
\multirow{4}{*}{ReAct-RAG}
  & Llama-3.1-8B  & 0.729 \\
  & Llama-3.1-70B & 0.679 \\
  & Qwen3-32B     & 0.813 \\
  & GPT-4o-mini   & 0.712 \\
\midrule
\textbf{\textsc{EviGen} (ours)} & --- & \textbf{0.871} \\
\bottomrule
\end{tabular}
\caption{Prediction performance of ReAct-RAG on the mortality task. \textsc{EviGen}'s prediction is produced by its Evidence Selection Layer and does not depend on the generation backbone.}
\label{tab:react-pred}
\end{table}

\begin{table}[t]
\centering
\footnotesize
\setlength{\tabcolsep}{4pt}
\begin{tabular}{l l r r r}
\toprule
\textbf{Backbone} & \textbf{Method} & S1 & S2 & S1$\And$S2 \\
\midrule
\multirow{2}{*}{Llama-3.1-8B}
  & ReAct-RAG & 8.5  & 69.3 & 5.9 \\
  & \textsc{EviGen}    & 72.6 & 81.0 & \textbf{58.8} \\
\midrule
\multirow{2}{*}{Llama-3.1-70B}
  & ReAct-RAG & 60.9 & 80.3 & 48.9 \\
  & \textsc{EviGen}    & 67.7 & 83.1 & \textbf{56.3} \\
\midrule
\multirow{2}{*}{Qwen3-32B}
  & ReAct-RAG & 23.5 & 94.6 & 22.2 \\
  & \textsc{EviGen}    & 55.7 & 94.2 & \textbf{52.5} \\
\midrule
\multirow{2}{*}{GPT-4o-mini}
  & ReAct-RAG & 29.7 & 94.5 & 28.1 \\
  & \textsc{EviGen}    & 72.9 & 88.6 & \textbf{64.6} \\
\bottomrule
\end{tabular}
\caption{Rationale faithfulness of ReAct-RAG and \textsc{EviGen} on the mortality task. S1 = source grounding pass rate; S2 = reasoning consistency pass rate (evaluated on factors passing S1); S1$\And$S2 = joint pass rate across all 22 checks (\S\ref{sec:eval:faithfulness}). The higher S1$\And$S2 per backbone is in \textbf{bold}.}
\label{tab:react-faith}
\end{table}

\section{Component-Analysis Results}
\label{app:component-full}

This appendix reports the full results for the component analysis in \S\ref{sec:component}: the matched $2\times2$ faithfulness results for all four backbones and conditions (Table~\ref{tab:component-2x2}), the marginal effect of each component (Table~\ref{tab:component-marginal}), and the evidence-ranking comparison (Table~\ref{tab:ranking-value}).

\begin{table}[t]
\centering
\footnotesize
\setlength{\tabcolsep}{3pt}
\begin{tabular}{l l r r r}
\toprule
\textbf{Backbone} & \textbf{Setting} & S1 & S2 & S1$\And$S2 \\
\midrule
\multirow{4}{*}{Llama-3.1-8B}
  & IG + scaffold   & 72.6 & 81.0 & \textbf{58.8} \\
  & RAG + scaffold  & 64.3 & 74.6 & 48.0 \\
  & IG + free-form  & 65.7 & 70.8 & 46.6 \\
  & RAG + free-form & 51.7 & 66.1 & 34.1 \\
\midrule
\multirow{4}{*}{Llama-3.1-70B}
  & IG + scaffold   & 67.7 & 83.1 & \textbf{56.3} \\
  & RAG + scaffold  & 62.1 & 81.1 & 50.4 \\
  & IG + free-form  & 45.8 & 80.1 & 36.7 \\
  & RAG + free-form & 36.4 & 75.8 & 27.6 \\
\midrule
\multirow{4}{*}{Qwen3-32B}
  & IG + scaffold   & 55.7 & 94.2 & \textbf{52.5} \\
  & RAG + scaffold  & 50.4 & 93.5 & 47.1 \\
  & IG + free-form  & 33.8 & 92.6 & 31.3 \\
  & RAG + free-form & 33.6 & 88.8 & 29.9 \\
\midrule
\multirow{4}{*}{GPT-4o-mini}
  & IG + scaffold   & 72.9 & 88.6 & \textbf{64.6} \\
  & RAG + scaffold  & 67.2 & 89.2 & 59.9 \\
  & IG + free-form  & 64.4 & 89.6 & 57.7 \\
  & RAG + free-form & 54.7 & 86.5 & 47.3 \\
\bottomrule
\end{tabular}
\caption{Matched $2\times2$ faithfulness comparison on the mortality task, crossing evidence source (IG-ranked vs.\ similarity-ranked RAG, top-5 chunks each) with generation input (\textsc{EviGen}'s evidence scaffold vs.\ free-form). ``IG + scaffold'' corresponds to \textsc{EviGen}. The RAG + free-form condition uses the matched top-5 budget and shared predicted probability, so its values differ from the full RAG baseline in Table~\ref{tab:faithfulness}. The highest S1$\And$S2 per backbone is in \textbf{bold}.}
\label{tab:component-2x2}
\end{table}

\begin{table}[t]
\centering
\footnotesize
\setlength{\tabcolsep}{2pt}
\begin{tabular}{l rrr rrr}
\toprule
 & \multicolumn{3}{c}{Scaffold} & \multicolumn{3}{c}{Evidence} \\
\cmidrule(lr){2-4}\cmidrule(lr){5-7}
\textbf{Backbone} & S1 & S2 & S1$\And$S2 & S1 & S2 & S1$\And$S2 \\
\midrule
Llama-8B    & +9.7  & +9.4 & +13.0 & +11.2 & +5.6 & +11.6 \\
Llama-70B   & +23.8 & +4.2 & +21.2 & +7.5  & +3.2 & +7.5  \\
Qwen3-32B   & +19.3 & +3.2 & +19.2 & +2.7  & +2.2 & +3.4  \\
GPT-4o-mini & +10.5 & +0.9 & +9.7  & +7.7  & +1.3 & +7.5  \\
\bottomrule
\end{tabular}
\caption{Marginal component effects from the matched $2\times2$ experiment (Table~\ref{tab:component-2x2}): average change in each faithfulness metric from adding the evidence scaffold (left) and from switching similarity-ranked to IG-ranked evidence (right).}
\label{tab:component-marginal}
\end{table}

\begin{table}[t]
\centering
\footnotesize
\setlength{\tabcolsep}{4pt}
\begin{tabular}{l r r}
\toprule
\textbf{Consumer} & \textbf{IG Top-5} & \textbf{Similarity Top-5} \\
\midrule
\textsc{EviGen} predictor & 0.868 & 0.661 \\
Qwen3-32B          & 0.803 & 0.589 \\
Llama-3.1-8B       & 0.742 & 0.596 \\
Llama-3.1-70B      & 0.737 & 0.599 \\
GPT-4o-mini        & 0.706 & 0.560 \\
\bottomrule
\end{tabular}
\caption{Prediction AUC on the mortality task when each consumer model predicts from a matched five-chunk evidence set, ranked either by \textsc{EviGen}'s Integrated Gradients attribution or by cosine similarity to the fixed RAG queries.}
\label{tab:ranking-value}
\end{table}

\section{Compute and Hardware}
\label{app:compute}

\textsc{EviGen}'s Evidence Selection Layer and \textsc{IRIS} are trained on a single A5000 Ada GPU (32~GB VRAM). LLM inference uses one H200 (141~GB VRAM) for 8B-scale models (Llama-3.1-8B-Instruct), two H200s in parallel for 32B and 70B models (Qwen3-32B-thinking and Llama-3.1-70B-Instruct), and four H200s for fine-tuning jobs (QLoRA baselines and the process-supervised verifier). GPT-4o-mini is accessed via the Azure OpenAI API.

\section{Datasets}
\label{app:datasets}

Tables~\ref{tab:dataset-cohort} and~\ref{tab:dataset-length} report cohort sizes and per-patient history-length statistics for the three datasets used in this paper. This appendix gives their cohort definitions and preprocessing; the train/validation/test splits and class balance are described in \S\ref{sec:experiments}.

\subsection{MIMIC-IV mortality}
\label{app:datasets:mimic}
We use MIMIC-IV \citep{johnson2023mimic} for the 1-year all-cause mortality task. For each patient, the index event is their last hospital discharge in the available record; the outcome is whether the patient dies within 365 days of that discharge, determined from the linked death-information table. MIMIC-IV is a de-identified clinical dataset with credentialed access requirements designed to protect patient privacy.

\paragraph{Inclusion criteria.}
\begin{itemize}[nosep,leftmargin=1.5em]
  \item Age $> 40$ years at the index discharge.
  \item At least 3 discharge notes across the patient's available history.
  \item Discharge density $> 1$ discharge per 200 days across the patient's available history, retaining patients with an active care trajectory rather than isolated visits.
\end{itemize}
For each included patient, the input to all methods is the concatenation of their discharge notes and structured ICD diagnostic codes up to and including the index discharge, ordered chronologically.

\subsection{Autism and ADHD early risk prediction}
\label{app:datasets:peds}
The autism and ADHD datasets are early-age risk prediction tasks drawn from an institutional pediatric EHR system. For each patient, the input is the EHR available before a fixed age cutoff (1.5 years for autism, 3 years for ADHD), and the outcome is whether the patient subsequently receives a clinical diagnosis of the target condition. Both cohorts share the same structure: (i) a positive cohort defined by a confirmed diagnosis plus an age-window visit requirement, and (ii) a negative cohort drawn from the same EHR system. Cohort sizes are reported in Table~\ref{tab:dataset-cohort}.

\paragraph{Autism cohort.} A patient is included as a positive if all of the following hold:
\begin{itemize}[nosep,leftmargin=1.5em]
  \item Born on or after 2014-01-01, so that the full early-childhood EHR window is captured by the source system.
  \item Has a recorded clinical diagnosis of autism spectrum disorder at any point in the available record.
  \item Has at least one well-child visit in each of three age windows: 0-6 months, 6-12 months, and 12-18 months, so that all positives have a comparable baseline of prediagnostic primary-care contact.
\end{itemize}
The model input for each autism patient is the EHR (clinical notes and ICD codes) accrued before age 1.5 years.

\paragraph{ADHD cohort.} A patient is included as a positive if all of the following hold:
\begin{itemize}[nosep,leftmargin=1.5em]
  \item Born on or after 2014-01-01.
  \item Has a recorded clinical diagnosis of ADHD at any point in the available record.
  \item Has at least one well-child visit in each of four age windows: 0-6 months, 6-12 months, 12-24 months, and 24-36 months.
\end{itemize}
The model input for each ADHD patient is the EHR accrued before age 3 years.

\paragraph{Leakage controls.} Label leakage is controlled in two ways. First, prediction uses only EHR data recorded before the age cutoff (1.5 years for autism, 3 years for ADHD), both well before the typical age of diagnosis, so information dated after the cutoff, including target-condition referrals, diagnostic mentions, and ICD codes, cannot enter the model input. Second, we exclude patients with autism-related or ADHD-related ICD-10 codes recorded before the corresponding cutoff.

\paragraph{Negative selection.} For each task, negatives are drawn from patients in the same EHR system who meet the same visit-window requirement as the corresponding positive cohort and have no recorded diagnosis of the target condition at any point in the available record. Negatives are frequency-matched to positives on the marginal distributions of age, sex, birth year, year of last encounter, and total observed EHR duration from the first to the last recorded encounter; duration matching reduces the risk of assigning negative labels simply because patients have shorter observation histories and fewer opportunities to receive a diagnosis. A negative label denotes no recorded target diagnosis in the available EHR, rather than confirmed lifetime absence of the condition, as matching cannot guarantee complete post-cutoff follow-up or capture diagnoses made outside our health system.

\paragraph{Institutional protocol.} EHR extraction, deidentification, and analysis were carried out under approval of the relevant institutional review board.

\subsection{Preprocessing}
\label{app:datasets:preprocessing}
For all three datasets, free-text notes are split into 200-token chunks with overlap and embedded with Qwen3-Embedding-8B (Appendix~\ref{app:a1:encoder}); each chunk is prefixed with an age-aware header so the embedding carries demographic context. ICD codes are embedded from their official long-title descriptions with the same age prefix and deduplicated at the (patient, code, version) level so a recurring diagnosis is counted once per patient.

\begin{table}[t]
\small
\centering
\setlength{\tabcolsep}{4pt}
\begin{tabular}{l r r r r}
\toprule
\textbf{Dataset} & \textit{Total} & \textit{Pos.} & \textit{Neg.} & \textit{Prev.} \\
\midrule
Mortality (MIMIC-IV) & 13{,}394 & 5{,}445 & 7{,}949 & 40.7\% \\
Autism               & 9{,}597  & 1{,}602 & 7{,}995 & 16.7\% \\
ADHD                 & 9{,}756  & 1{,}767 & 7{,}989 & 18.1\% \\
\bottomrule
\end{tabular}
\caption{Cohort sizes and class balance for the three datasets.}
\label{tab:dataset-cohort}
\end{table}

\begin{table*}[t]
\small
\centering
\setlength{\tabcolsep}{5pt}
\begin{tabular}{l r r r r r r r r}
\toprule
\textbf{Dataset} & \textit{min} & \textit{p25} & \textit{median} & \textit{p75} & \textit{p90} & \textit{p99} & \textit{max} & \textit{mean} \\
\midrule
Mortality (MIMIC-IV) & 1{,}946 & 7{,}000  & 9{,}972  & 16{,}190 & 28{,}252  & 78{,}507      & 266{,}860     & 14{,}691  \\
Autism               & 4{,}502 & 33{,}599 & 49{,}011 & 72{,}146 & 119{,}421 & 1{,}411{,}827 & 5{,}055{,}035 & 100{,}964 \\
ADHD                 & 3{,}292 & 41{,}622 & 58{,}735 & 86{,}242 & 138{,}778 & 1{,}392{,}111 & 8{,}587{,}998 & 108{,}465 \\
\bottomrule
\end{tabular}
\caption{Per-patient history length in tokens, using the Llama-3.1-8B tokenizer on the raw patient text without truncation. Autism and ADHD statistics are computed on the training split; validation and test splits show comparable distributions. The long right tail (p99 in the millions) reflects a small subset of patients with extensive prior healthcare contact. The 122{,}880-token input budget used by LLM baselines (Table~\ref{tab:baseline-config}) is exceeded by 9.5\% of autism and 12.7\% of ADHD test patients, who are left-truncated for the baselines; \textsc{EviGen}'s retrieval-based design processes all patients without truncation.}
\label{tab:dataset-length}
\end{table*}

\section{Faithfulness Evaluation Methodology}
\label{app:faithfulness}

\paragraph{Note-only setup.} Faithfulness is evaluated on note-only rationales: every generated rationale is required to cite clinical notes as evidence, not ICD codes. ICD code descriptions are short and easy to reproduce verbatim, while note chunks are long and information-dense and much more prone to paraphrasing or hallucination when quoted. Because LLMs differ in how often they prefer ICD versus note evidence, allowing both types would conflate a model's quote fidelity with its preference for easy-to-quote material. By restricting evidence to notes, we put all models in the same quoting setting, so S1 (and therefore S1$\And$S2) reflects faithfulness rather than evidence-type preference.

\subsection{Stage 1: source grounding}
\label{app:faithfulness:stage1}

\paragraph{String-match algorithm.} For each cited quote in a rationale, we score how well it matches the cited source note using \texttt{rapidfuzz.partial\_ratio}. Given two strings $s_1, s_2$, let $\text{short}$ be the shorter of the two and let $w$ range over substrings of the longer string with length $|\text{short}|$. The score is
\[
\text{partial\_ratio}(s_1, s_2) = 100 \cdot \max_{w}\, \frac{\mathrm{LCS}(\text{short}, w)}{|\text{short}|},
\]
where $\mathrm{LCS}(\cdot, \cdot)$ denotes the longest common subsequence length. Intuitively, the shorter string is slid across the longer one and the best-aligned window's LCS-based similarity is reported. We treat a quote as grounded if the score is $\geq 90$, which tolerates roughly 5-10\% character-level disagreement, typically from minor punctuation or whitespace artifacts.

\paragraph{Text normalization.} Both the source text (when the note index is built) and the rationale quote (at match time) are passed through the same normalization pipeline, so that the comparison happens in a single canonical form: (i) Unicode NFKC normalization, which collapses visually equivalent but distinct code points (fullwidth digits, ligatures, etc.); (ii) replacement of non-breaking spaces (U+00A0) with regular spaces, which NFKC does not handle; (iii) collapsing of consecutive whitespace into a single space, since clinical notes are heavily multiline while LLM-generated quotes are typically single-line; (iv) trimming and lowercasing. Without this step, the partial-ratio score is depressed by formatting differences that carry no clinical content.

\paragraph{Ellipsis-segmented quotes.} Some reasoning-model backbones (notably Qwen3-32B in thinking mode) occasionally write quotes containing an ellipsis (e.g., ``A \ldots B'') to indicate that part of the original passage has been omitted. A direct partial-ratio against the source then scores poorly because the source contains no contiguous ``A\ldots B''. We split such quotes on the ellipsis marker, score each segment against the source independently, and take the minimum segment score as the quote's grounding score. A quote with skipped content is therefore accepted only if every contiguous segment is individually grounded.

\subsection{Stage 2: reasoning consistency}
\label{app:faithfulness:stage2}

\paragraph{Judge model and setup.} We use GPT-4o (via the Azure OpenAI API) as the Stage 2 judge. The judge sees only the rationale being audited; the patient EHR is not provided. This keeps the per-rationale judging cost low (one short call regardless of EHR length) and forces the judge to assess internal coherence between each factor's quoted observation and its stated reasoning, rather than re-deriving conclusions from the full record.

\paragraph{Prompt and output schema.} Table~\ref{tab:judge-prompt} gives the verbatim system prompt, the user message structure, and the JSON schema the judge is asked to return. The schema produces two booleans per factor (\texttt{reasoning\_follows\_from\_evidence}, \texttt{reasoning\_uses\_only\_stated\_evidence}) and two overall booleans (\texttt{overall\_reasoning\_follows\_from\_factors}, \texttt{overall\_uses\_only\_stated\_factors}); field definitions are listed in the same table.

\paragraph{Aggregation.} The judge produces two booleans per factor and two overall booleans, giving $2 \times 5 + 2 = 12$ Stage 2 checks per rationale. Combined with the two Stage 1 checks per factor (citation ID validity, evidence matching), each 5-factor rationale has $2_{\text{S1}} \times 5 + 2_{\text{S2}} \times 5 + 2_{\text{overall}} = 22$ checks. A rationale passes Stage 2 only if all 12 Stage 2 booleans are true; the S1$\And$S2 pass rate reported in \S\ref{sec:results:faithfulness} requires all 22.

\section{Clinical Reviewer Pilot}
\label{app:reviewer-pilot}

\paragraph{Study design.} We recruited 7 medical students to evaluate three rationale types: (i)~the \textsc{EviGen} rationale (prediction + evidence-scaffolded reasoning + process verification), (ii)~a standard LLM free-form explanation (prediction + unstructured narrative), and (iii)~the raw output of \textsc{EviGen}'s Evidence Selection Layer (prediction probability + top-attribution chunks with scores, no LLM narrative). Each of the three systems generated rationales for the same 20 randomly selected patients, and every reviewer reviewed the rationales from all three systems, presented in randomized order. Reviewers were given only the generated rationales, with no access to the underlying EHR or any other patient information; they therefore assumed predictions were correct and cited quotes were not hallucinated. This design is deliberately conservative toward \textsc{EviGen}: holding prediction correctness and quote fidelity fixed across formats isolates explanation quality and removes the two axes on which \textsc{EviGen} is measurably stronger (\S\ref{sec:results:prediction}, \S\ref{sec:results:faithfulness}). Each reviewer rated all three rationale types on a 30-item Likert questionnaire (1 = strongly disagree, 5 = strongly agree) and provided an overall preference ranking. Four items are reverse-coded; all scores below are reported after reverse-coding. This study was approved by the Institutional Review Board (IRB) of our institution. All participants provided informed consent prior to participation.

\paragraph{Constructs.} The 30 items are organized into five constructs drawn from validated user-evaluation scales: \textbf{Usability} (14 items from SUS~\citep{brooke1996sus}, HSUS~\citep{ghorayeb2023design}, and SCS~\citep{holzinger2020measuring}), \textbf{Informativeness} (3 items from HSUS and ESS~\citep{hoffman2023measures}), \textbf{Actionability} (6 items from HSUS), \textbf{Trust} (6 items from SUS, HSUS, and Hoffman Trust~\citep{hoffman2023measures}), and \textbf{Engagement} (1 item from SUS).

\subsection{Construct-level ratings}
\label{app:reviewer-results}
\begin{figure}[t]
  \centering
  \includegraphics[width=\columnwidth]{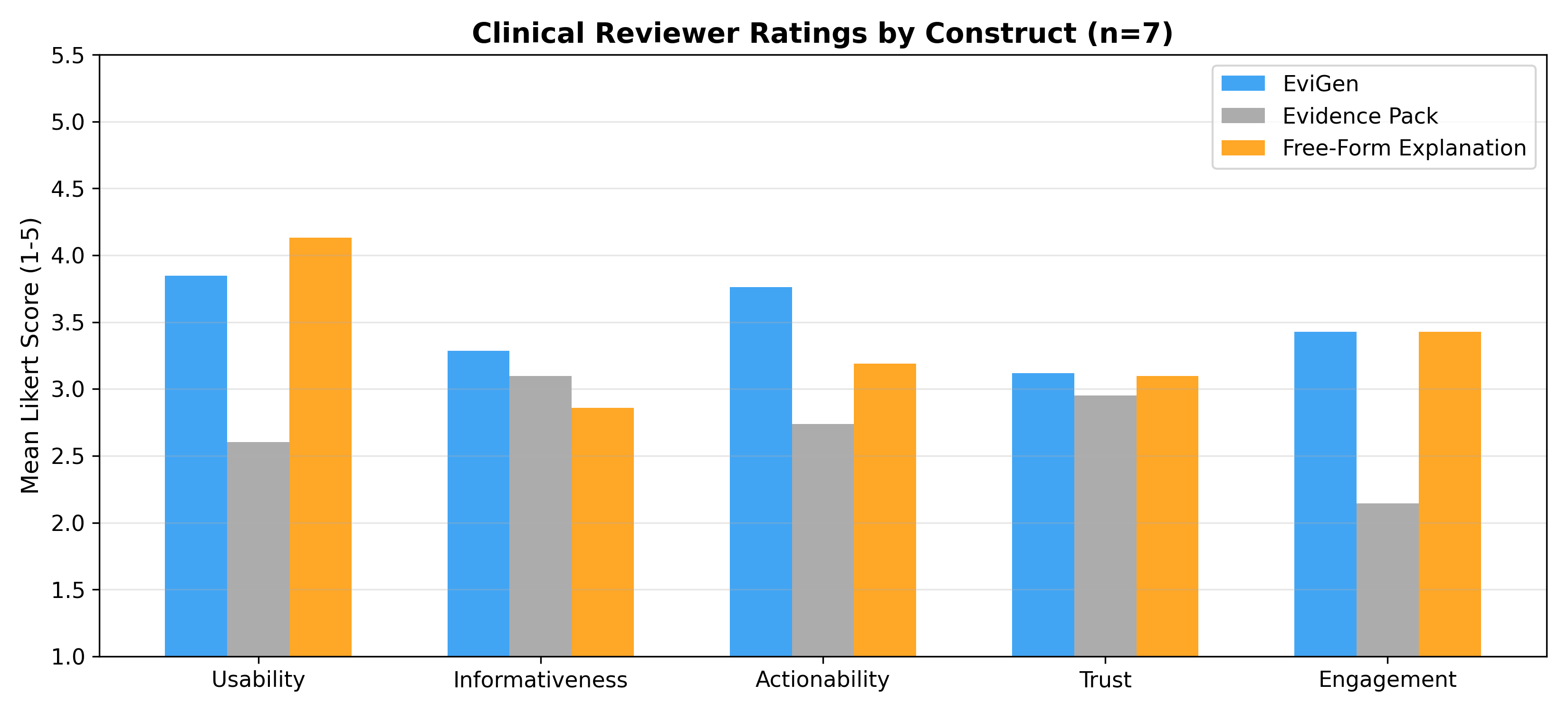}
  \caption{Construct-level mean Likert ratings (1-5) from the clinical reviewer pilot ($n=7$ medical students). Each construct aggregates a subset of the survey items in Table~\ref{tab:clinical-survey}. \textsc{EviGen} leads on Informativeness, Actionability, and Trust; the free-form LLM explanation rates slightly higher on Usability.}
  \label{fig:clinical-ratings}
\end{figure}

Figure~\ref{fig:clinical-ratings} reports the mean Likert score (1-5) per evaluation construct, averaged across the 7 reviewers for each of the three reviewed methods: \textsc{EviGen}, the standard LLM free-form explanation, and the raw Evidence Selection Layer output (``Evidence Pack''). The 30 survey items in Table~\ref{tab:clinical-survey} are aggregated into five constructs (Usability, Informativeness, Actionability, Trust, Engagement) by averaging the items assigned to each construct. \textsc{EviGen} leads on Informativeness, Actionability, and Trust; the free-form LLM explanation edges \textsc{EviGen} on Usability (simpler output, no attribution scores or verifier flags to interpret); the bare Evidence Pack trails both narrative formats on most constructs and notably on Engagement. Mean (SD) values per construct are reported in Table~\ref{tab:construct-scores}; on the standardized SUS usability scale (0-100), scores are 70.0 (``OK'') for \textsc{EviGen}, 83.6 (``Good'') for Free-Form, and 30.7 (``Poor'') for Evidence Pack.

\begin{table}[t]
\centering
\small
\setlength{\tabcolsep}{3pt}
\begin{tabular}{l r@{\hskip 3pt}l r@{\hskip 3pt}l r@{\hskip 3pt}l}
\toprule
 & \multicolumn{2}{c}{\textsc{EviGen}} & \multicolumn{2}{c}{Evid.\ Pack} & \multicolumn{2}{c}{Free-Form} \\
\textbf{Construct} & \multicolumn{2}{c}{\textit{M (SD)}} & \multicolumn{2}{c}{\textit{M (SD)}} & \multicolumn{2}{c}{\textit{M (SD)}} \\
\midrule
Usability (14)       & 3.85 & (0.41) & 2.60 & (0.61) & \textbf{4.13} & (0.46) \\
Informativeness (3)  & \textbf{3.29} & (0.45) & 3.10 & (0.90) & 2.86 & (0.42) \\
Actionability (6)    & \textbf{3.76} & (0.53) & 2.74 & (0.63) & 3.19 & (0.44) \\
Trust (6)            & \textbf{3.12} & (0.51) & 2.95 & (0.36) & 3.10 & (0.25) \\
Engagement (1)       & 3.43 & (0.98) & 2.14 & (1.07) & 3.43 & (0.54) \\
\addlinespace
\midrule
Overall              & \textbf{3.49} & & 2.71 & & 3.34 & \\
\bottomrule
\end{tabular}
\caption{Construct-level clinical reviewer ratings (mean and SD across $n{=}7$ reviewers, 1-5 Likert scale). Number in parentheses after construct name is item count. Best per row in \textbf{bold}.}
\label{tab:construct-scores}
\end{table}

\subsection{Per-item results}
\label{app:per-item-results}
Table~\ref{tab:per-item-scores} reports the mean score for each of the 30 survey items across the three systems. \textsc{EviGen} shows its largest advantages on items related to supporting (not dictating) decisions (Q23: 4.43 vs.\ 3.43), improving care quality (Q19: 3.71 vs.\ 3.00), and showing sufficient detail (Q16: 3.29 vs.\ 2.00). Free-Form Explanation leads on pure usability items such as complexity (Q1: 5.00 vs.\ 4.00) and navigability (Q8: 4.43 vs.\ 3.57).

\subsection{Survey Questionnaire}
\label{app:clinical-survey}

Each reviewer rated 30 items per rationale on a 5-point Likert scale (1 = Strongly Disagree, 5 = Strongly Agree). The items are adapted from four validated user-evaluation scales for decision-support systems and grouped by source scale in Table~\ref{tab:clinical-survey}. The placeholder \texttt{[System]} stands for the method being rated, and is substituted at survey time with \textsc{EviGen}, the standard LLM rationale (prediction + free-form explanation), or the raw output of \textsc{EviGen}'s Evidence Selection Layer.

\begin{table*}[p]
\small
\centering
\begin{tabular}{>{\raggedright\arraybackslash}p{0.18\textwidth} >{\raggedright\arraybackslash}p{0.75\textwidth}}
\toprule
\textbf{Field} & \textbf{Content} \\
\midrule
Role & Clinical interpretability assistant for \textsc{EviGen}, a model that predicts patient outcomes from structured and unstructured EHR evidence. \\
\addlinespace
Task & Binary prediction of all-cause mortality within 1 year after the patient's last hospital discharge. \\
\addlinespace
Input & (1) Predicted probability; (2) top-$k$ predictive factors (ICD codes and note snippets) with signed attribution score and similarity score. \\
\addlinespace
Goals & Conservative, evidence-grounded interpretation. Do not infer diagnoses, severity, or mechanisms beyond what is explicitly stated. Keep physiologic reasoning general; provide only high-level care considerations. \\
\midrule
\multicolumn{2}{l}{\textbf{Output structure (in this exact order)}} \\
\midrule
{}[Prediction + Explanation] & State that this is a prediction of all-cause mortality within 1 year; report a population percentile if provided; summarize main clinical patterns in plain language. Indicate chronic/acute/end-stage only if clear from the evidence. \\
\addlinespace
{}[Evidence Reasoning] & For each factor, use the four labeled fields below. \\
\addlinespace
Factor summary & Strictly descriptive paraphrase of the evidence or ICD label; no added diagnoses, mechanisms, or severity descriptors. \\
\addlinespace
Supporting evidence & Short VERBATIM quote from the note or ICD description, with documented date/time for notes. Every line MUST end with a citation in one of: \texttt{(note\_id: XXXXX)} or \texttt{(ICD code: XXXXX-X)}, copied verbatim. \\
\addlinespace
Reasoning & 1-2 plain sentences: restate the clinical finding, give a broad physiologic/prognostic implication, then connect to 1-year mortality risk. Stay general and conservative. For healthcare-process documentation (e.g., teach-back attestations, template language), frame as a healthcare-engagement proxy and stop there. The conclusion direction must match the attribution sign (positive $\Rightarrow$ pushes risk up; negative $\Rightarrow$ pushes risk down). \\
\addlinespace
Attribution score & Line of the form \texttt{Attribution score: X}, where X is the signed attribution to 4 decimal places (e.g., \texttt{+1.2744} or \texttt{-0.0786}). This is a feature-contribution value, not a probability change. \\
\addlinespace
{}[Reasoning Lens] & 3-4 sentences synthesizing the factors into a coherent narrative for the predicted risk. Do not introduce new diagnoses or events; describe temporal patterns only if implied by the evidence. \\
\addlinespace
{}[Recommendations] & General considerations (prognosis communication, goals-of-care, symptom-focused support) and one tentative consideration per factor. Tentative language only (``may'', ``could be relevant''); no specific diagnostics, medications, or procedures. \\
\addlinespace
Closing rule & If uncertain whether a statement is supported by the evidence, omit it. \\
\bottomrule
\end{tabular}
\caption{Prompt structure for the rationale-generation LLM (mortality task variant). The 4-section output schema and citation format are enforced by the prompt. Supporting-evidence lines must end with one of two citation formats, which the citation rule in \S\ref{sec:method:generation} mechanically verifies.}
\label{tab:prompt}
\end{table*}

\begin{figure*}[p]
  \centering
  \includegraphics[width=\textwidth]{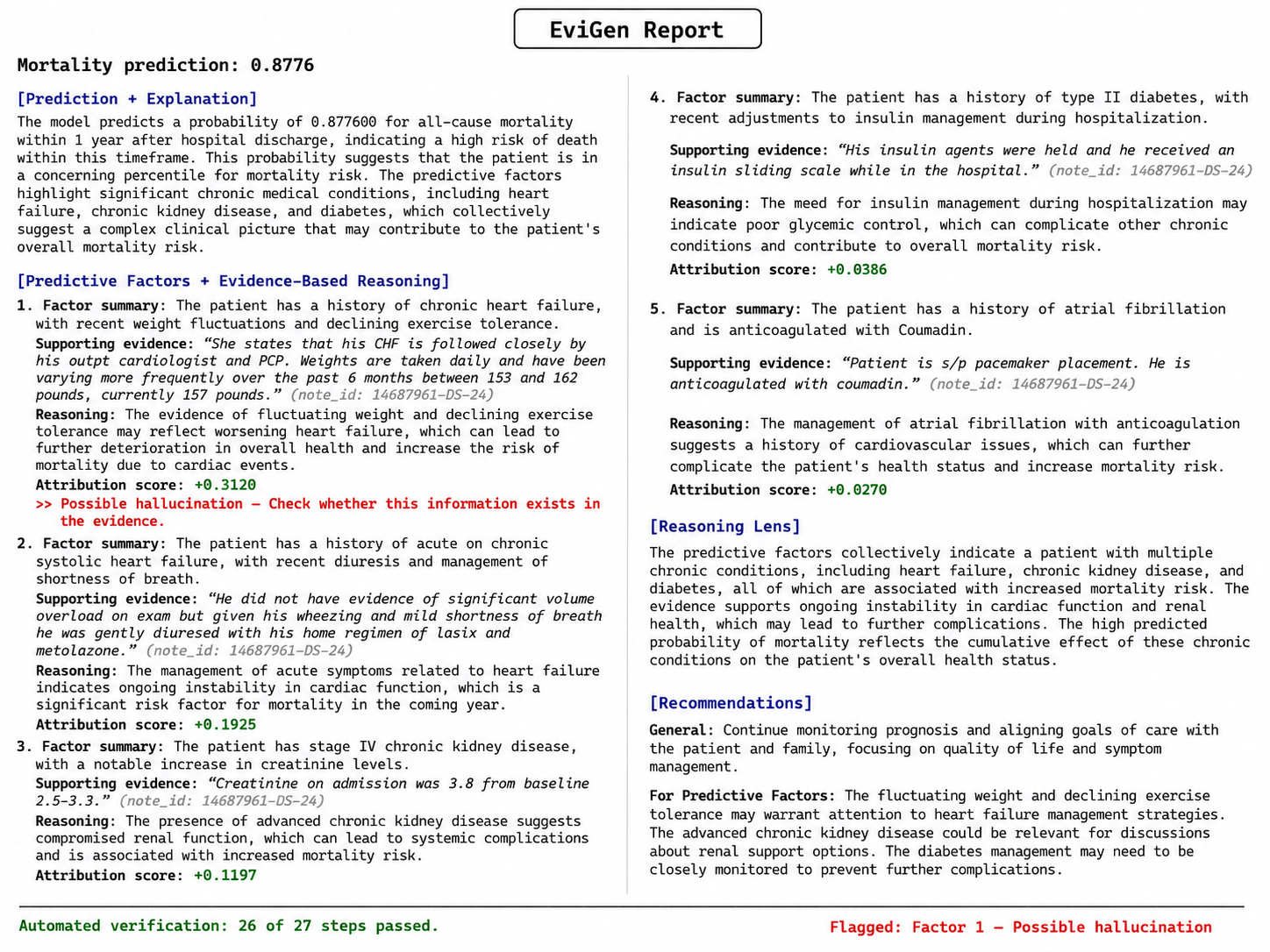}
  \caption{Complete \textsc{EviGen}-generated rationale (mortality task) with verifier step-level verification overlay. The verifier flags Factor 1 as a possible hallucination: the cited quote supports the weight fluctuation but does not mention the ``declining exercise tolerance'' that appears in the factor summary and reasoning. The bottom status line summarizes per-step verification results.}
  \label{fig:flagged-example}
\end{figure*}

\begin{table*}[p]
\small
\centering
\begin{tabular}{>{\raggedright\arraybackslash}p{0.22\textwidth} >{\raggedright\arraybackslash}p{0.72\textwidth}}
\toprule
\textbf{Field} & \textbf{Content} \\
\midrule
\multicolumn{2}{l}{\textbf{System prompt}} \\
\midrule
Role and goal & You are a clinical interpretability assistant. Given a patient's complete clinical history (clinical notes and ICD diagnostic codes in chronological order), you (1) estimate the probability of all-cause mortality within 1 year of the last hospital discharge, and (2) produce a conservative, evidence-grounded clinical rationale identifying the top 5 predictive factors drawn ONLY from the provided history. \\
\addlinespace
Hard rules & Do NOT infer diagnoses, severity, or mechanisms beyond what is explicitly stated in the evidence. Keep physiologic reasoning general and non-specific. Provide only high-level care considerations, not specific treatment plans. \\
\addlinespace
Citation format & Cite EVERY supporting-evidence line using one of these EXACT formats, copying the identifier verbatim from the patient history: \texttt{(note\_id: XXXXX)} for clinical-note evidence; \texttt{(ICD code: XXXXX-X)} for ICD-based evidence. Do not use any other citation format. Omit attribution scores. \\
\addlinespace
Output discipline & Respond with STRICT JSON only. \\
\midrule
\multicolumn{2}{l}{\textbf{User instruction}} \\
\midrule
Intro & Below is a single patient's complete clinical history, ordered from oldest to newest. Each note header includes its \texttt{note\_id} and each code line includes its ICD code-version identifier. \\
\addlinespace
Step 1 & Estimate the probability (float in $[0.0, 1.0]$) that this patient will die within one year after their last discharge. \\
\addlinespace
Step 2 & Write a clinical rationale following exactly the structure below; output it as the value of the \texttt{report} field (use \texttt{\textbackslash n} for line breaks). \\
\addlinespace
{}[Prediction + Explanation] & State that this is a prediction of all-cause mortality within 1 year; report the estimated probability; summarize main clinical patterns in plain, non-technical language. Mention \emph{chronic}, \emph{acute}, or \emph{end-stage} only if clearly supported. Related factors may be grouped under a brief, clinically intuitive heading when supported. \\
\addlinespace
{}[Predictive Factors + Evidence-Based Reasoning] & For each of the top 5 factors: (i) \emph{Factor summary}: strictly descriptive paraphrase of the evidence or ICD label; no added diagnoses, mechanisms, or severity. (ii) \emph{Supporting evidence}: short direct quote (notes) or ICD description, with documented Date and Age for notes; every line ends with a citation in the exact format above. (iii) \emph{Reasoning}: one sentence in the form \textit{(evidence as stated)} $\rightarrow$ \textit{(broad physiologic or prognostic implication)} $\rightarrow$ \textit{(impact on 1-year mortality risk)}. \\
\addlinespace
{}[Reasoning Lens] & 3-4 sentences synthesizing the 5 factors into a coherent narrative for the estimated risk. No new diagnoses or events. \\
\addlinespace
{}[Recommendations] & 1-2 general considerations (prognosis communication, goals-of-care, symptom-focused support) plus one tentative care consideration per factor. Tentative language only (``may'', ``could be relevant''); no specific diagnostics, medications, or procedures. \\
\addlinespace
Output schema & Respond with ONLY a single JSON object, no prose outside the object, no markdown fences: \texttt{\{"probability": <float in [0.0, 1.0]>, "report": <string>\}}. \\
\bottomrule
\end{tabular}
\caption{Verbatim full-context zero-shot baseline prompt (mortality task variant). The output schema mirrors \textsc{EviGen}'s rationale structure (Figure~\ref{fig:report-example}) but omits the attribution-score field.}
\label{tab:zeroshot-prompt}
\end{table*}

\begin{table*}[p]
\small
\centering
\begin{tabular}{>{\raggedright\arraybackslash}p{0.32\textwidth} >{\raggedright\arraybackslash}p{0.61\textwidth}}
\toprule
\textbf{Field} & \textbf{Content} \\
\midrule
\multicolumn{2}{l}{\textbf{System prompt}} \\
\midrule
\multicolumn{2}{p{0.94\textwidth}}{\itshape You are auditing a clinical rationale for internal coherence. You will see ONE rationale. The rationale contains 5 numbered factors; each factor has a Supporting evidence line and a Reasoning sentence. The rationale ends with a Reasoning Lens that synthesizes the 5 factors. Do NOT judge medical correctness or whether the prognosis is right. Judge only whether the reasoning the rationale itself states follows from the evidence the rationale itself stated. Return strict JSON.} \\
\midrule
\multicolumn{2}{l}{\textbf{User message}} \\
\midrule
Body & The full text of the rationale under audit, followed by the schema instruction. \\
\addlinespace
Schema instruction & \textit{Return a JSON object with this exact shape (no extra keys):} (template below) \\
\midrule
\multicolumn{2}{l}{\textbf{JSON output schema}} \\
\midrule
\multicolumn{2}{l}{\texttt{\{}} \\
\multicolumn{2}{l}{\texttt{~~"per\_factor": [}} \\
\multicolumn{2}{l}{\texttt{~~~~\{"idx": <int 1-5>,}} \\
\multicolumn{2}{l}{\texttt{~~~~~"reasoning\_follows\_from\_evidence": <bool>,}} \\
\multicolumn{2}{l}{\texttt{~~~~~"reasoning\_uses\_only\_stated\_evidence": <bool>,}} \\
\multicolumn{2}{l}{\texttt{~~~~~"rationale": "<=1 short sentence"\}}} \\
\multicolumn{2}{l}{\texttt{~~],}} \\
\multicolumn{2}{l}{\texttt{~~"overall\_reasoning\_follows\_from\_factors": <bool>,}} \\
\multicolumn{2}{l}{\texttt{~~"overall\_uses\_only\_stated\_factors": <bool>,}} \\
\multicolumn{2}{l}{\texttt{~~"overall\_rationale": "<=2 sentences"}} \\
\multicolumn{2}{l}{\texttt{\}}} \\
\midrule
\multicolumn{2}{l}{\textbf{Field definitions}} \\
\midrule
\texttt{reasoning\_\allowbreak follows\_\allowbreak from\_\allowbreak evidence}        & Does the factor's Reasoning sentence logically follow from its Supporting evidence? \\
\addlinespace
\texttt{reasoning\_\allowbreak uses\_\allowbreak only\_\allowbreak stated\_\allowbreak evidence}    & Does the factor avoid introducing facts not present in the Supporting evidence (e.g., diagnoses, dates, severity not stated in the quote)? \\
\addlinespace
\texttt{overall\_\allowbreak reasoning\_\allowbreak follows\_\allowbreak from\_\allowbreak factors} & Does the Reasoning Lens follow from the 5 factors? \\
\addlinespace
\texttt{overall\_\allowbreak uses\_\allowbreak only\_\allowbreak stated\_\allowbreak factors}       & Does the Reasoning Lens avoid introducing factors or evidence not listed above? \\
\bottomrule
\end{tabular}
\caption{Stage 2 reasoning-consistency judge prompt and JSON output schema. The judge (GPT-4o) sees only the rationale under audit; the patient EHR is not passed in. A rationale passes Stage 2 only if all per-factor booleans are \texttt{true} for every factor and both overall booleans are \texttt{true}.}
\label{tab:judge-prompt}
\end{table*}

\begin{table*}[p]
\small
\centering
\begin{tabular}{c >{\raggedright\arraybackslash}p{0.85\textwidth}}
\toprule
\# & \textbf{Item} \\
\midrule
\multicolumn{2}{l}{\textbf{System Usability Scale (SUS; \citealp{brooke1996sus})}} \\
\midrule
1 & I found [System] unnecessarily complex. \\
2 & I thought [System] was easy to use. \\
3 & I would imagine that most people would learn to use [System] very quickly. \\
4 & I found [System] very cumbersome to use. \\
5 & I felt very confident using [System]. \\
6 & I thought there was too much inconsistency in [System]. \\
7 & I think that I would like to use [System] frequently. \\
\midrule
\multicolumn{2}{l}{\textbf{Health-IT System Usability Scale (HSUS; \citealp{ghorayeb2023design})}} \\
\midrule
8  & [System] fits well with the way I currently work. \\
9  & I found the information provided on the screen understandable. \\
10 & I found it easy to navigate through [System]. \\
11 & The screen layout makes it easy to see each piece of information. \\
12 & On the screen, I can find specific information I need quickly. \\
13 & [System] generates a useful summary view of the patient's current health status. \\
14 & [System] helps me work more efficiently. \\
15 & I am able to provide better quality of care for patients by using [System]. \\
16 & It is easier to make efficient decisions by using [System]. \\
17 & [System] helps improve patient outcomes. \\
18 & [System] helps prevent clinical errors. \\
\midrule
\multicolumn{2}{l}{\textbf{System Causability Scale (SCS; \citealp{holzinger2020measuring})}} \\
\midrule
19 & I understood the explanations within the context of my work. \\
20 & I did not need support to understand the explanations. \\
21 & I was able to use the explanations with my knowledge base. \\
22 & I think that most people would learn to understand the explanations very quickly. \\
23 & [System]'s explanation has sufficient detail. \\
\midrule
\multicolumn{2}{l}{\textbf{Explanation Satisfaction \& Trust (ESS / Trust; \citealp{hoffman2023measures})}} \\
\midrule
24 & [System]'s explanation shows me how accurate [System] is. \\
25 & [System] supports my decision-making rather than dictating it. \\
26 & I understand how [System] creates its recommendations, scores, or alerts. \\
27 & [System]'s recommendations, scores, or alerts are consistent with clinical practices and standards. \\
28 & I believe the recommendations, scores, or alerts are reliable. \\
29 & I am confident in [System]. I feel that it works well. \\
30 & I am wary of [System]. \\
\bottomrule
\end{tabular}
\caption{Clinical reviewer survey items, grouped by source scale. Each reviewer rated each item on a 5-point Likert scale (1 = Strongly Disagree, 5 = Strongly Agree). The placeholder \texttt{[System]} is substituted with the method being rated (\textsc{EviGen}, the standard LLM rationale, or the raw Evidence Selection Layer output).}
\label{tab:clinical-survey}
\end{table*}

\begin{table*}[p]
\centering
\footnotesize
\setlength{\tabcolsep}{4pt}
\begin{tabular}{c l l r r r}
\toprule
\textbf{\#} & \textbf{Source} & \textbf{Item} & \textbf{\textsc{EviGen}} & \textbf{Evid.\ Pack} & \textbf{Free-Form} \\
\midrule
\multicolumn{6}{l}{\emph{Usability}} \\
Q1  & SUS     & Unnecessarily complex (R)              & 4.00 & 2.14 & \textbf{5.00} \\
Q2  & SUS     & Easy to use                             & 3.86 & 2.00 & \textbf{4.43} \\
Q3  & SUS     & Learn quickly                           & 3.71 & 2.29 & \textbf{4.57} \\
Q4  & SUS     & Cumbersome (R)                          & 4.14 & 2.29 & \textbf{4.71} \\
Q5  & SUS     & Confident using                         & \textbf{3.29} & 2.43 & 3.00 \\
Q6  & HSUS    & Fits workflow                           & 3.43 & 2.29 & \textbf{3.57} \\
Q7  & HSUS    & Screen understandable                   & 4.00 & 3.00 & \textbf{4.43} \\
Q8  & HSUS    & Easy to navigate                        & 3.57 & 2.29 & \textbf{4.43} \\
Q9  & HSUS    & Layout clear                            & 3.43 & 2.29 & \textbf{4.00} \\
Q10 & HSUS    & Find info quickly                       & 3.43 & 2.00 & \textbf{3.57} \\
Q11 & SCS     & Explanations in context                 & \textbf{4.43} & 3.57 & 4.14 \\
Q12 & SCS     & No support needed                       & \textbf{4.00} & 3.43 & \textbf{4.00} \\
Q13 & SCS     & Use with knowledge base                 & \textbf{4.43} & 3.57 & 4.14 \\
Q14 & SCS     & Learn explanations quickly              & \textbf{4.14} & 2.86 & 3.86 \\
\cmidrule(lr){1-6}
\multicolumn{6}{l}{\emph{Informativeness}} \\
Q15 & HSUS    & Useful summary view                     & 3.71 & 2.57 & \textbf{4.00} \\
Q16 & ESS     & Sufficient detail                       & 3.29 & \textbf{3.43} & 2.00 \\
Q17 & ESS     & Shows accuracy                          & 2.86 & \textbf{3.29} & 2.57 \\
\cmidrule(lr){1-6}
\multicolumn{6}{l}{\emph{Actionability}} \\
Q18 & HSUS    & Work efficiently                        & \textbf{4.00} & 2.29 & 3.86 \\
Q19 & HSUS    & Better quality care                     & \textbf{3.71} & 2.57 & 3.00 \\
Q20 & HSUS    & Easier decisions                        & \textbf{4.00} & 2.57 & 3.00 \\
Q21 & HSUS    & Improve outcomes                        & \textbf{3.29} & 3.00 & 3.14 \\
Q22 & HSUS    & Prevent errors                          & \textbf{3.14} & 3.00 & 2.71 \\
Q23 & HSUS    & Supports not dictates                   & \textbf{4.43} & 3.00 & 3.43 \\
\cmidrule(lr){1-6}
\multicolumn{6}{l}{\emph{Trust}} \\
Q24 & SUS     & Too much inconsistency (R)              & 3.57 & 3.86 & \textbf{4.00} \\
Q25 & HSUS    & Understand how it works                 & \textbf{2.71} & 2.14 & 2.43 \\
Q26 & HSUS    & Consistent with standards               & 3.14 & \textbf{3.43} & 3.29 \\
Q27 & HSUS    & Reliable                                & 3.00 & \textbf{3.43} & \textbf{3.43} \\
Q28 & Hoffman & Confident in system                     & \textbf{3.29} & 2.43 & 3.00 \\
Q29 & Hoffman & Wary (R)                                & \textbf{3.00} & 2.43 & 2.43 \\
\cmidrule(lr){1-6}
\multicolumn{6}{l}{\emph{Engagement}} \\
Q30 & SUS     & Use frequently                          & \textbf{3.43} & 2.14 & \textbf{3.43} \\
\bottomrule
\end{tabular}
\caption{Per-item mean scores across $n{=}7$ reviewers (1-5 Likert, after reverse-coding). Items marked (R) are reverse-coded. Best per row in \textbf{bold}.}
\label{tab:per-item-scores}
\end{table*}

\end{document}